\documentclass[sigconf,anonymous=false]{acmart}

\usepackage{amsmath,color,boldline}
\usepackage{multirow}
\usepackage[linesnumbered,ruled,vlined]{algorithm2e}
\usepackage{algpseudocode}
\usepackage{enumerate}

\usepackage{xcolor}
\usepackage{graphicx}
\usepackage{appendix}
\usepackage{tabularx}
\usepackage{subcaption}
\usepackage{paralist}
\usepackage{lipsum}
\usepackage{makecell}
\usepackage{enumitem}
\usepackage{footmisc}
\usepackage{wrapfig}

\setitemize{noitemsep,topsep=0pt,parsep=0pt,partopsep=0pt}

\newtheorem{problem}{Problem}

\newtheorem{lemma}{Lemma}

\algnewcommand\algorithmicinput{\textbf{Input:}}
\algnewcommand\INPUT{\item[\algorithmicinput]}
\algnewcommand\algorithmicoutput{\textbf{Output:}}
\algnewcommand\OUTPUT{\item[\algorithmicoutput]}

\newcommand{\hide}[1]{}

\newcommand{\gasoline}{\textsc{GaSoliNe}}
\newcommand{\nettack}{\textsc{Nettack}}

\AtBeginDocument{%
  \providecommand\BibTeX{{%
    \normalfont B\kern-0.5em{\scshape i\kern-0.25em b}\kern-0.8em\TeX}}}

\setcopyright{none}
\copyrightyear{}
\acmYear{}
\acmDOI{}
\renewcommand\footnotetextcopyrightpermission[1]{}

\begin{document}

\title{Graph Sanitation with Application to Node Classification}

\author{Zhe Xu}
\email{zhexu3@illinois.edu}
\affiliation{
  \institution{University of Illinois at Urbana-Champaign}
}

\author{Boxin Du}
\email{boxindu2@illinois.edu}
\affiliation{
  \institution{University of Illinois at Urbana-Champaign}
}

\author{Hanghang Tong}
\email{htong@illinois.edu}
\affiliation{
  \institution{University of Illinois at Urbana-Champaign}
}

\begin{abstract}



The past decades have witnessed the prosperity of graph mining, with a multitude of sophisticated models and algorithms designed for various mining tasks, such as ranking, classification, clustering and anomaly detection. Generally speaking, the vast majority of the existing works aim to answer the following question, that is, {\em given a graph, what is the best way to mine it?}

In this paper, we introduce the {\em graph sanitation} problem, to answer an orthogonal question. That is, {\em given a mining task and an initial graph, what is the best way to improve the initially provided graph?} By learning a better graph as part of the input of the mining model, it is expected to benefit graph mining in a variety of settings, ranging from denoising, imputation to defense. We formulate the graph sanitation problem as a bilevel optimization problem, and further instantiate it by semi-supervised node classification, together with an effective solver named \gasoline. Extensive experimental results demonstrate that the proposed method is (1) broadly applicable with respect to various graph neural network models and flexible graph modification strategies, (2) effective in improving the node classification accuracy on both the original and contaminated graphs in various perturbation scenarios. In particular, it brings up to 25\% performance improvement over the existing robust graph neural network methods.

\end{abstract}


\maketitle

\section{Introduction}
\label{sec:introduction}


Graph mining has become the cornerstone in a wealth of real-world applications, such as social media mining~\cite{zafarani2014social}, brain connectivity analysis~\cite{alper2013weighted}, computational epidemiology~\cite{keeling2005networks} and financial fraud detection~\cite{wang2019semi}. For the vast majority of existing works, they essentially aim to answer the following question, that is, \emph{given} a graph, what is the best model and/or algorithm to mine it? To name a few, PageRank~\cite{page1999pagerank} and its variants~\cite{jeh2003scaling,tong2006fast,haveliwala2003topic,gyongyi2004combating} measure the node importance and node proximity based on multiple weighted paths; spectral clustering~\cite{shi2000normalized} minimizes inter-cluster connectivity and maximizes the intra-cluster connectivity to partition nodes into different groups; graph neural networks (GNNs)~\cite{kipf2016semi,velivckovic2018graph, wu2019simplifying, klicpera2018predict} learn representation of nodes by aggregating information from the neighborhood. 
In all these works and many more, they require a {\em given} graph, including its topology and/or the associated attribute information, as part of the input of the corresponding mining model.

Despite tremendous success, some fundamental questions largely remain open, e.g., where does the input graph come from at the first place? To what extent does the quality of the given graph impact the effectiveness of the corresponding graph mining model? 
In response, we introduce the \emph{graph sanitation} problem, which aims to improve the initially provided graph for a given graph mining model, so as to maximally boost its performance. The rationality is as follows. In many existing graph mining works, the initially provided graph is typically constructed manually based on some heuristics. The graph construction is often treated as a pre-processing step, without the consideration of the specific mining task. What is more, the initially constructed graph could be subject to various forms of contamination, such as missing information, noise and even adversarial attacks. This suggests that there might be under-explored space for improving the mining performance by learning a `better' graph as the input of the corresponding task.

There are a few lines of existing works for modifying graphs. 
For example, network imputation~\cite{liben2007link,huisman2009imputation} and knowledge graph completion~\cite{bordes2013translating,wang2014knowledge} problems focus on restoring missing links in a partially observed graph; graph connectivity optimization~\cite{chen2018network} and computational immunization~\cite{chen2015node} problems aim to manipulate the graph connectivity in a desired way by changing the underlying topology; robust GNNs~\cite{entezari2020all,wu2019adversarial,DBLP:conf/kdd/Jin0LTWT20} utilize empirical properties of a benign graph to remove or assign lower weights to the poisoned graph elements (e.g., contaminated edges). 


The graph sanitation problem introduced in this paper is related to but bears subtle difference from these existing work in the following sense. Most, if not all, of these existing works for modifying graphs assume the initially provided graph is impaired or perturbed in a specific way, e.g., due to missing links, or noise, or adversarial attacks. Some existing works further impose certain assumptions on the specific graph modification algorithms, such as the low-rank assumption behind many network imputation methods, the types of attacks and/or the empirical properties of the benign graph (e.g., topology sparsity, feature smoothness) behind some robust GNNs. In contrast, the proposed graph sanitation problem does not make any such assumptions, but instead pursues a different design principle. That is, we aim to let the performance of the downstream data mining task, measured on a validation set, dictate how we should optimally modify the initially provided graph. This is crucial, as it not only ensures that the modified graph will directly and maximally improve the mining performance, but also lends itself to be applicable to a variety of graph mining tasks. 

Formally, we formulate the graph sanitation problem as a generic bilevel optimization problem, where the lower-level optimization problem corresponds to the specific mining task and the upper-level optimization problem encodes the supervision to modify the provided graph and maximally improve the mining performance. 
Based on that, we instantiate such a bilevel optimization problem by semi-supervised node classification with GNNs, where the lower-level objective function represents the cross-entropy classification loss over the training data and the upper-level objective function represents the loss over validation data, using the mining model trained from the lower-level optimization problem. We propose an effective solver (\gasoline) which adopts an efficient approximation of hyper-gradient to guide the modification over the given graph. We carefully design the hyper-gradient aggregation mechanism to avoid potential bias from a specific dataset split by aggregating the hyper-gradient from different folds of data. 
\gasoline\ is versatile, and is equipped with multiple variants, such as discretized vs. continuous modification, modifying graph topology vs. feature.
Comprehensive experiments demonstrate that (1) \gasoline\ is broadly applicable to benefit different downstream node classifiers together with flexible choices of variants and modification strategies, (2) \gasoline~ can significantly boost downstream classifiers on both the original and contaminated graphs in various perturbation scenarios and can work hand-in-hand with existing robust GNNs methods. For instance, in Table~\ref{tab:improve existing methods}, 
the proposed \gasoline\ significantly boosts GAT~\cite{velivckovic2018graph}, SVD~\cite{entezari2020all}, and RGCN~\cite{zhu2019robust}.

\begin{table}[t]
\resizebox{0.9\linewidth}{!}{
\begin{tabular}{cc|ccc}
\toprule
\textbf{Data}             & \textbf{With \gasoline?} & \textbf{GAT} & \textbf{SVD} & \textbf{RGCN} \\ \midrule
\multirow{2}{*}{Cora}     & N                   & 48.8$\pm$0.2   & 60.1$\pm$0.6       & 50.6$\pm$0.8    \\
                          & Y                   & 63.7$\pm$0.6   & 79.7$\pm$0.6       & 62.6$\pm$0.6    \\ \midrule
\multirow{2}{*}{Citeseer} & N                   & 62.4$\pm$0.7   & 50.6$\pm$0.6       & 55.5$\pm$1.4    \\
                          & Y                   & 69.7$\pm$0.2   & 76.5$\pm$0.6       & 66.1$\pm$0.8    \\ \midrule
\multirow{2}{*}{Polblogs} & N                   & 48.2$\pm$6.6   & 77.3$\pm$3.3       & 50.8$\pm$0.9    \\
                          & Y                   & 70.8$\pm$0.6   & 89.2$\pm$0.7       & 67.7$\pm$0.3    \\ \bottomrule
\end{tabular}}
\caption{Node classification performance (Mean$\pm$Std Accuracy) boosting of existing defense methods on heavily poisoned graphs ($25\%$ edges perturbed by metattack~\cite{DBLP:conf/iclr/ZugnerG19}) by the proposed \gasoline.}
\vspace{-6mm}
\label{tab:improve existing methods}
\end{table}

In summary, our main contributions in this paper are as follows:
\begin{itemize}[noitemsep,leftmargin=9pt]
    \item \textbf{Problem Definition.} We introduce a novel graph sanitation problem, and formulate it as a bilevel optimization problem. The proposed graph sanitation problem can be potentially applied to a variety of graph mining models as long as they are differentiable w.r.t. the input graph.
    \item \textbf{Algorithmic Instantiation.} We instantiate the graph sanitation problem by semi-supervised node classification with GNNs. We further propose an effective and scalable solver named \gasoline\ with versatile variants. 
    \item \textbf{Empirical Evaluations.} We perform extensive empirical studies on real-world datasets to demonstrate the effectiveness and the applicability 
    of the proposed \gasoline\ algorithms. 
\end{itemize}

\section{Graph Sanitation Problem}
\label{sec:problemDefinition}


\noindent{\bf A - Notations.} We use bold uppercase letters for matrices (e.g., $\mathbf{A}$), bold lowercase letters for column vectors (e.g., $\mathbf{u}$), lowercase letters for scalars (e.g., $c$), and calligraphic letters for sets (e.g., $\mathcal{T}$). We use $\mathbf{A}[i,j]$ to represent the entry of matrix $\mathbf{A}$ at the $i$-th row and the $j$-th column, $\mathbf{A}[i,:]$ to represent the $i$-th row of matrix $\mathbf{A}$, and $\mathbf{A}[:,j]$ to represent the $j$-th column of matrix $\mathbf{A}$. Similarly, $\mathbf{u}[i]$ denotes the $i$-th entry of vector $\mathbf{u}$. We use prime to denote the transpose of matrices and vectors (e.g., $\mathbf{A}^{\prime}$ is the transpose of $\mathbf{A}$). For the variables of the modified graphs, we set $\tilde{\:}$ over the corresponding variables of the original graphs (e.g., $\tilde{\mathbf{A}}$).  



We represent an attributed graph as  $G=\{\mathbf{A},\mathbf{X}\}$, where $\mathbf{A}\in\mathbb{R}^{n\times n}$ is the adjacency matrix and  $\mathbf{X}\in\mathbb{R}^{n\times d}$ is the feature matrix composed by $d$-dimensional feature vectors of $n$ nodes. For \emph{supervised graph mining models}, we first divide the node set into two disjoint subsets: labeled node set $\mathcal{Z}$ and test set $\mathcal{W}$, and then divide the labeled node set $\mathcal{Z}$ into two disjoint subsets: the training set $\mathcal{T}$ and the validation set $\mathcal V$. We use $y$ and $\hat{y}$ with appropriate indexing to denote the ground truth supervision and the prediction result respectively. Take the classification task as an example, $y_{ij}=1$ if node $i$ belongs to class $j$ and $y_{ij}=0$ otherwise; $\hat{y}_{ij}$ is the predicted probability that node $i$ belongs to class $j$. Furthermore, we use $\mathcal{Y}_{\texttt{train}}$ and $\mathcal{Y}_\texttt{valid}$ to denote the supervision information of all the nodes in the training set $\mathcal{T}$ and the validation set $\mathcal{V}$, respectively.

\noindent {\bf B - Optimization-Based Graph Mining Models.}
For many graph mining models, they can be formulated from the optimization perspective~\cite{kang2019n2n} with a general goal to find an optimal solution $\theta^*$ so that a task-specific loss $\mathcal{L}(G,\theta, \mathcal{T}, \mathcal{Y}_\texttt{train})$ is minimized. Here, $\mathcal{T}$ and $\mathcal{Y}_\texttt{train}$ are the training set and the associated ground truth (e.g., class labels for the classification task), which would be absent for the unsupervised graph mining tasks (e.g., clustering, ranking). We give three concrete examples next. 


\begin{table*}[th]
\resizebox{\linewidth}{!}{
\begin{tabular}{c|c|c|c}
\hline
\textbf{Mining Tasks} & Personalized PageRank~\cite{jeh2003scaling,backstrom2011supervised,li2016quint} & Spectral clustering~\cite{shi2000normalized, wang2010flexible} & Semi-supervised node classification \\ \hline
$\mathcal{L}_{\texttt{lower}}$          & $\min_{\mathbf{r}}\  q\mathbf{r}^{\prime}(\mathbf{I}-\bar{\mathbf{A}})\mathbf{r}+(1-q)||\mathbf{r}-\mathbf{e}||^2$ & \makecell{$\min_{\mathbf{u}}\ \mathbf{u}^{\prime}\mathbf{L}\mathbf{u}\quad
    \textrm{s.t.} \ \mathbf{u}^{\prime}\mathbf{D}\mathbf{u}=1, \  \mathbf{D}\mathbf{u}\perp\mathbf{1}$} & $\min_\theta\  -\sum_{i \in \mathcal{T}}\sum_{j=1}^c y_{ij} \ln{\hat{y}_{ij}}$ \\ \hline
$\mathcal{L}_{\texttt{upper}}$          & $\min_{\mathbf{A}}\ \sum_{x\in\mathcal{P},y\in\mathcal{N}}(1+\exp{(\mathbf{r}^*[x]-\mathbf{r}^*[y])/w})^{-1}$ & $\min_{\mathbf{A}}\ -\mathbf{u}^{*\prime}\mathbf{Q}\mathbf{u}^*$ & $\min_G\  -\sum_{i \in \mathcal{V}}\sum_{j=1}^c y_{ij} \ln{\hat{y}_{ij}}$ \\ \hline
$\mathcal{T}$                & none & none & training set $\mathcal{T}$ \\ \hline
$\mathcal{Y}_{\texttt{train}}$             & none & none & labels of training set $\mathcal{Y}_{\texttt{train}}$ \\ \hline
$\mathcal{V}$                & \makecell{positive node set $\mathcal{P}$\\negative node set $\mathcal{N}$} & \makecell{`must-link' set $\mathcal{M}$ \\   `cannot-link' set $\mathcal{C}$} & validation set $\mathcal{V}$ \\ \hline
$\mathcal{Y}_\texttt{valid}$             & none & none & labels of validation set $\mathcal{Y}_{\texttt{valid}}$ \\ \hline
\textbf{Remarks} & \makecell{normalized adjacency matrix $\bar{\mathbf{A}}$\\damping factor $q$ \\ preference vector $\mathbf{e}$ \\ width parameter $w$} & \makecell{Laplacian matrix $\mathbf{L}$\\degree matrix $\mathbf{D}$\\link constraints matrix $\mathbf{Q}$} & \makecell{number of classes $c$ \\ predicted probability of node $i$ to class $j$ $\hat{y}_{ij}$ \\ binary ground truth of node $i$ to class $j$ $y_{ij}$} \\ \hline
\end{tabular}
}
\caption{Instantiations of graph sanitation problem over various mining tasks}
\vspace{-9mm}
\label{tab:instantiations}
\end{table*}

{\em Example \#1: personalized PageRank}~\cite{jeh2003scaling} is a fundamental ranking model. When the adjacency matrix of the underlying graph is symmetrically normalized, the ranking vector $\mathbf r$ can be obtained as:

\vspace{-3mm}
\begin{equation}
    \mathbf{r}^*=\arg\min_{\mathbf{r}}\  q\mathbf{r}^{\prime}(\mathbf{I}-\bar{\mathbf{A}})\mathbf{r}+(1-q)||\mathbf{r}-\mathbf{e}||^2,
    \label{eq:pagerank loss}
\end{equation}
\vspace{-4mm}

\noindent where $\bar{\mathbf{A}}$ is the symmetrically normalized adjacency matrix; $q\in(0,1]$ is the damping factor; $\mathbf{e}$ is the preference vector; the ranking vector $\mathbf{r}^*$ is the solution of the ranking model (i.e., $\theta^*=\mathbf{r}^*$).

{\em Example \#2: spectral clustering}~\cite{shi2000normalized} is a classic graph clustering model aiming to minimizes the normalized cut between clusters:

\vspace{-4mm}
\begin{equation}
\begin{split}
    \mathbf{u}^*= \arg\min_{\mathbf{u}}\ \mathbf{u}^{\prime}\mathbf{L}\mathbf{u}\quad
    \textrm{s.t.} \  \mathbf{u}^{\prime}\mathbf{D}\mathbf{u}=1,\  \mathbf{D}\mathbf{u}\perp\mathbf{1}_\mathbf{u},
    \label{eq:spectral clustering loss}
\end{split}
\end{equation}
\vspace{-5mm}

\noindent where $\mathbf{L}$ is the Laplacian matrix of adjacency matrix, $\mathbf{D}$ is the diagonal degree matrix (i.e., $\mathbf{D}[i,i]=\sum_{j}\mathbf{A}[i,j]$), $\mathbf{1}_u$ is an all-one vector with the same size as $\mathbf{u}$; the model solution $\theta^*$ is the cluster indicator vector $\mathbf{u}^*$.

{\em Example \#3: node classification} aims to construct a classification model based on the graph topology $\mathbf A$ and feature $\mathbf X$. A typical loss for node classification is cross-entropy (CE) over the training set: 
\vspace{-1mm}
\begin{equation}
\begin{split}
\theta^* &= \arg\min_{\theta}\ -\sum_{i \in \mathcal{T}}\sum_{j=1}^c y_{ij} \ln{\hat{y}_{ij}},
\end{split}
\label{eq:node classification loss}
\end{equation}
\vspace{-3mm}

\noindent where $c$ is the number of classes, $y_{ij}$ is the ground truth indicating if node $i$ belongs to class $j$, $\mathcal{T}$ is the training set, $\hat{y}_{ij} = f(G, \theta)[i,j]$ is the predicted probability that node $i$ belongs to class $j$ by a classifier $f(G, \theta)$ parameterized by $\theta$. For example, classifier $f(G, \theta)$ can be a GNN whose trained model parameters form the solution $\theta^*$

{\em Remarks.} Both the standard personalized PageRank and spectral clustering are unsupervised and therefore the training set $\mathcal T$ and its supervision $\mathcal{Y}_\texttt{train}$ are absent in the corresponding loss functions (i.e., Eq.~\eqref{eq:pagerank loss} and \eqref{eq:spectral clustering loss}, respectively). Nonetheless, both personalized PageRank and spectral clustering have been generalized to further incorporate some forms of supervision, as we will show next.

\noindent{\bf C - Graph Sanitation: Formulation and Instantiations.} Given an initial graph $G$ and an optimization-based graph mining model $\mathcal{L}(G,\theta, \mathcal{T}, \mathcal{Y}_\texttt{train})$, we aim to learn a modified graph $\tilde{G}$ to boost the performance of the corresponding mining model and we name it as \emph{graph sanitation problem}. The basic idea is to let the mining performance on a validation set $\mathcal{V}$ guide the modification process. Formally, the graph sanitation problem is defined as follows. 


\vspace{-2mm}
\begin{problem}{Graph Sanitation Problem}
\vspace{-1mm}
\label{prof:graph sanitation problem}
\begin{description}
\item[Given:] (1) a graph represented as  $G=\{\mathbf{A},\mathbf{X}\}$, (2) a graph mining task represented as $\mathcal{L}(G,\theta, \mathcal{T}, \mathcal{Y}_{\texttt{train}})$, (3) a validation set $\mathcal{V}$ and its supervision $\mathcal{Y}_\texttt{valid}$, and (4) the sanitation budget $B$;
\item[Find:] A modified graph $\tilde{G}=\{\tilde{\mathbf{A}},\tilde{\mathbf{X}}\}$ to boost the performance of input graph mining model.
\end{description}
\end{problem}
\vspace{-2mm}

We formulate Problem~\ref{prof:graph sanitation problem} as a bilevel optimization problem:

\vspace{-4mm}
\begin{equation}\label{eq:bilevel:opt}
\begin{split}
    \tilde{G}=&\arg\min_{G}\ \mathcal{L}_{\texttt{upper}}(G,\theta^*,\mathcal{V},\mathcal{Y}_\texttt{valid})\\
    \textrm{s.t.} &\  \theta^* = \arg\min\mathcal{L}_{\texttt{lower}}(G,\theta,\mathcal{T},\mathcal{Y}_{\texttt{train}}),\ D(\tilde{G},G)\leq B\\
\end{split}
\end{equation}
\vspace{-3mm}


\noindent where the lower-level optimization is to train the model $\theta^*$ based on the training set $\mathcal T$; the upper-level optimization aims to optimize the performance of the trained model $\theta^*$ on the validation set $\mathcal V$, and there is no overlap between $\mathcal T$ and $\mathcal V$; the distance function $D$ measures the distance between two graphs. For example, we can instantiate $D(\tilde{G},G)$ as $||\tilde{\mathbf{A}}-\mathbf{A}||_{1,1}$ or $||\tilde{\mathbf{X}}-\mathbf{X}||_{1,1}$ based on scenarios.
Notice that the loss function at the upper level $\mathcal{L}_{\texttt{upper}}$ might be different from the one at the lower level $\mathcal{L}_{\texttt{lower}}$. For example,  $\mathcal{L}_{\texttt{lower}}$ for both personalized PageRank (Eq.~\eqref{eq:pagerank loss}) and spectral clustering (Eq.~\eqref{eq:spectral clustering loss}) does not involve any supervision. However,  $\mathcal{L}_{\texttt{upper}}$ for both models is designed to measure the performance on a validation set with supervision and therefore should be different from $\mathcal{L}_{\texttt{lower}}$. We elaborate this next. 

The proposed bilevel optimization problem in Eq.~\eqref{eq:bilevel:opt} is quite general. In principle, it is applicable to {\em any} graph model with differentiable $\mathcal{L}_{\texttt{upper}}$ and $\mathcal{L}_{\texttt{lower}}$. We give its instantiations with the three aforementioned mining tasks and summarize them in Table~\ref{tab:instantiations}.

{\em Instantiation \#1: supervised PageRank.} The original personalized PageRank~\cite{jeh2003scaling} has been generalized to encode pair-wised ranking preference~\cite{backstrom2011supervised, li2016quint}. For graph sanitation with supervised PageRank, the training set and its supervision is absent, and the lower-level loss $\mathcal{L}_{\texttt{lower}}$ is given in Eq.~\eqref{eq:pagerank loss}. The validation set $\mathcal V$ is consisted of a positive node set $\mathcal{P}$ and a negative node set $\mathcal{N}$. The supervision of the upper-level problem is that ranking scores of nodes from $\mathcal{P}$ should be higher than that from $\mathcal{N}$, i.e., $\mathbf{r}[x]>\mathbf{r}[y],\forall x\in\mathcal{P}, \forall y\in\mathcal{N}$. Several choices for the upper-level loss $\mathcal{L}_{\texttt{upper}}$ exist. For example, we can use Wilcoxon-Mann-Whitney loss~\cite{yan2003optimizing}:



\vspace{-3mm}
\begin{equation}
    \min_{\mathbf{A}}\quad\sum_{x\in\mathcal{P},y\in\mathcal{N}}\Big{(}1+\exp{(\mathbf{r}^*[x]-\mathbf{r}^*[y])/w}\Big{)}^{-1}
    \label{eq:graph sanitation for pagerank}
\end{equation}
\vspace{-2mm}

\noindent where $w$ is the width parameter. It is worth-mentioning that Eq.~\eqref{eq:graph sanitation for pagerank} only modifies graph topology $\mathbf{A}$. Although Eq.~\eqref{eq:graph sanitation for pagerank} does not contain variable $\mathbf{A}$, $\mathbf{r}^*$ is determined by $\mathbf{A}$ through the lower-level problem. 

{\em Instantiation \#2: supervised spectral clustering.} A typical way to encode supervision in spectral clustering is via
`must-link' and `cannot-link'~\cite{wagstaff2000clustering,wang2010flexible}. For graph sanitation with supervised spectral clustering, the training set together with its supervision is absent, and the lower-level loss $\mathcal{L}_{\texttt{lower}}$ is given in Eq.~\eqref{eq:spectral clustering loss}. The validation set $\mathcal V$ contains a `must-link' set $\mathcal{M}$ and a  `cannot-link' set $\mathcal{C}$. For the upper-level loss, the idea is to encourage nodes from must-link set $\mathcal M$ to be grouped in the same cluster and in the meanwhile push nodes from cannot-link set $\mathcal C$ to be in different clusters. To be specific, $\mathcal{L}_{\texttt{upper}}$ can be instantiated as follows.


\vspace{-5mm}
\begin{equation}
\begin{split}
    \min_{\mathbf{A}}\quad-\mathbf{u}^{*\prime}\mathbf{Q}\mathbf{u}^*
\end{split}
\label{eq:graph sanitation for spectral clustering}
\end{equation}
\vspace{-5mm}

\noindent where $\mathbf{Q}$ encodes the `must-link' and `cannot-link', that is, $\mathbf{Q}[i,j]=1$ if $(i,j)\in\mathcal{M}$, $\mathbf{Q}[i,j]=-1$ if $(i,j)\in\mathcal{C}$, and $\mathbf{Q}[i,j]=0$ otherwise. This instantiation only modifies the graph topology $\mathbf{A}$. 

{\em Instantiation \#3: semi-supervised node classification.} For graph sanitation with semi-supervised node classification, its lower-level optimization problem is given in Eq.~\eqref{eq:node classification loss}. We have cross-entropy loss over validation set $\mathcal{V}$ as the upper-level problem:

\vspace{-4mm}
\begin{equation}
    \min_{G}\quad\mathcal{L}_{\texttt{CE}}(G,\theta^*, \mathcal{V}, \mathcal{Y}_\texttt{valid})=-\sum_{i \in \mathcal{V}}\sum_{j=1}^c y_{ij} \ln{\hat{y}_{ij}}
    \label{eq:graph sanitation for semi-supervised node classification}
\end{equation}
\vspace{-3mm}

\noindent As mentioned before, there should be no overlap between the training set $\mathcal T$ and the validation set $\mathcal V$. If both the topology $\mathbf{A}$ and node feature $\mathbf{X}$ are used for classification, then both components can be modified in this instantiation.




{\em Remarks.} If the initially given graph $G$ is poisoned by adversarial attackers~\cite{zugner2018adversarial,DBLP:conf/iclr/ZugnerG19}, the graph sanitation problem with semi-supervised node classification can also be used as a defense strategy. However, it bears important difference from the existing robust GNNs~\cite{DBLP:conf/kdd/Jin0LTWT20,entezari2020all,wu2019adversarial} as it does not assume the given graph $G$ is poisoned or any specific way by which it is poisoned. Therefore, graph sanitation problem in this scenario can boost the performance under a wide range of attacking scenarios (e.g., non-poisoned graphs, lightly-poisoned graphs, and heavily-poisoned graphs) and has the potential to work hand-in-hand with existing robust GNNs model.
In the next section, we propose an effective algorithm to solve the graph sanitation problem with semi-supervised node classification.

\section{Proposed Algorithms: \gasoline}

\label{sec:method}

In this section, we focus on graph sanitation problem in the context of semi-supervised node classification and propose an effective solver named \gasoline. The general workflow of \gasoline\ is as follows. First, we solve the lower-level problem (Eq.~\eqref{eq:node classification loss}) and obtain a solution $\theta^*$ together with its corresponding updating trajectory. Then we compute the hyper-gradient of the upper-level loss function (Eq.~\eqref{eq:graph sanitation for semi-supervised node classification}) w.r.t. the graph $G$ and use a set of hyper-gradient-guided modification to solve the upper-level optimization problem. Recall that we need a classifier $f$ to provide the predicted labels (in both the lower-level and upper-level problems) which is parameterized by $\theta$ and we refer to this classifier as the \emph{backbone classifier}. Finally we test the performance of another classifier over the modified graph on the test set $\mathcal{W}$ and this classifier is named as the \emph{downstream classifier}. In the following subsections, we will introduce our proposed solution \gasoline\ in three parts, including (A) hyper-gradient computation, (B) hyper-gradient aggregation, (C) hyper-gradient-guided modification, and (D) low-rank speedup.

\noindent \textbf{A - Hyper-Gradient Computation.}
Eq.~\eqref{eq:bilevel:opt} and its corresponding instantiations Eqs.~\eqref{eq:node classification loss}\eqref{eq:graph sanitation for semi-supervised node classification} fall into the family of bilevel optimization problem where the lower-level problem is to optimize $\theta$ via minimizing the loss over the training set $\{\mathcal{T}, \mathcal{Y}_{\texttt{train}}\}$ given $G$, and the upper-level problem is to optimize $G$ via minimizing the loss over $\{\mathcal{V},\mathcal{Y}_{\texttt{valid}}\}$. We compute gradient w.r.t. the upper-level problem and view the lower-level problem as a dynamic system:
\vspace{-1mm}
\begin{equation}
    \theta^{t+1} = \Theta^{t+1}(G,\theta^{t},\mathcal{T}, \mathcal{Y}_{\texttt{train}}), \quad \theta^{1} = \Theta^{1}(G, \mathcal{T}, \mathcal{Y}_{\texttt{train}}),
    \label{eq:update theta}
\end{equation}
\vspace{-3mm}

\noindent where $\Theta^1$ is the initialization of $\theta$ and $\Theta^{t+1}$ ($t\neq0$) is the updating formula which can be instantiated as an optimizer over the lower-level objective function on training set (Eq.~\eqref{eq:node classification loss}). Hence, in order to get the hyper-gradient of the upper-level problem $\nabla_{G}\mathcal{L}$, we assume that the dynamic system converges in $T$ iterations (i.e., $\theta^* = \theta^{T}$). Then we can unroll the iterative solution of the lower-level problem and obtain the hyper-gradient $\nabla_{G}\mathcal{L}$ by the chain rule as follows~\cite{baydin2017automatic}. For brevity, we abbreviate cross-entropy loss over the validation set $\mathcal{L}_{\texttt{CE}}(G,\theta^T, \mathcal{V}, \mathcal{Y}_\texttt{valid})$ as $\mathcal{L}_{\texttt{valid}}(\theta^T)$.
\vspace{-1mm}
\begin{equation}
    \nabla_{G}\mathcal{L} = \nabla_{G}\mathcal{L}_{\texttt{valid}}(\theta^{T}) + \sum_{t=0}^{T-2}B_{t}A_{t+1}\dots A_{T-1}\nabla_{\theta^{T}}\mathcal{L}_{\texttt{valid}}(\theta^{T})
    \label{eq:hypergradient}
\end{equation}
\vspace{-1mm}

\noindent where $A_t=\nabla_{\theta^{t}}\theta^{t+1}$,  $B_t=\nabla_{G}\theta^{t+1}$.

Our final goal is to improve the performance of converged downstream classifiers. Hence, $T$ is set as a relatively large value (e.g., 200 in our experiments) to ensure the hyper-gradient from the upper-level problem is computed over a converged classifier. To balance the effectiveness and the efficiency, we adopt the truncated hyper-gradient~\cite{shaban2019truncated} w.r.t. $G$ and rewrite the second part of Eq.~\eqref{eq:hypergradient} as $\sum_{t=P}^{T-2}B_{t}A_{t+1}\dots A_{T-1}\nabla_{\theta^{T}}\mathcal{L}_{\texttt{valid}}(f_{\theta^{T}})$,
\noindent where $P$ denotes the truncating iteration. In order to achieve a faster estimation of the hyper-gradient, we further adopt a first-order approximation~\cite{nichol2018first,DBLP:conf/iclr/ZugnerG19} and the hyper-gradient $\nabla_{G}\mathcal{L}$ can be computed as:


\vspace{-3mm}
\begin{equation}
    \nabla_{G}\mathcal{L} = \sum_{t=P}^{T}{\nabla_{G}\mathcal{L}_{\texttt{valid}}({\theta}^{t})}.
    \label{eq:hyper-gradient approximation}
\end{equation}
\vspace{-3mm}

\noindent where the updating trajectory of $\theta^t$ is the same as Eq.~\eqref{eq:update theta}. If the initially-provided graph $G$ is undirected, it indicates that $\mathbf{A}=\mathbf{A}^{\prime}$. Hence, when we compute the hyper-gradient w.r.t. the undirected graph topology $\mathbf{A}$, we need to calibrate the partial derivative into the derivative~\cite{kang2019n2n} and update the hyper-gradient as follows:
\begin{equation}
    \nabla_{\mathbf{A}}\mathcal{L}\leftarrow\nabla_{\mathbf{A}}\mathcal{L}+(\nabla_{\mathbf{A}}\mathcal{L})^{\prime}-\textrm{diag}(\nabla_{\mathbf{A}}\mathcal{L}).
    \label{eq:fix derivative w.r.t. A}
\end{equation}
\noindent For the hyper-gradient w.r.t. feature $\mathbf{X}$ and directed graph topology $\mathbf{A}$ ($\mathbf{A}\neq\mathbf{A}^{\prime}$), the above calibration process is not needed.

\noindent \textbf{B - Hyper-Gradient Aggregation.}
To ensure the quality of graph sanitation without introducing bias from a specific dataset split, we adopt $K$-fold training/validation split with similar settings as cross-validation~\cite{chen2020trading}. The specific procedure is that during the training period, we split all the labeled nodes $\mathcal{Z}$ into $K$ folds and alternatively select one of them as $\mathcal{V}$ (with labels $\mathcal{Y}_{\texttt{valid}}$) and the others as $\mathcal{T}$ (with labels $\mathcal{Y}_{\texttt{train}}$). In total, there are $K$ sets of training/validation splits. With the $k$-th dataset split, by Eq.~\eqref{eq:hyper-gradient approximation}, we obtain the hyper-gradient $\nabla_{G}^k\mathcal{L}$. For the hyper-gradient $\{\nabla_{G}^{1}\mathcal{L},\dots,\nabla_{G}^{K}\mathcal{L}\}$ from the $K$ sets of training/validation split, we sum them up as the aggregated hyper-gradient: $\nabla_{G}= \sum_k{\nabla_{G}^{k}}\mathcal{L}$. 

\noindent \textbf{C - Hyper-Gradient-Guided Modification.}
With regard to modifying the graph based on the aggregated hyper-gradient $\nabla_{G}$, we provide two variants, \emph{discretized modification} and \emph{continuous modification}. The discretized modification can work with binary inputs such as adjacency matrices of unweighted graphs and binary feature matrices. The continuous modification is suitable for both continuous and binary inputs. For the clarity of explanation, we replace the $G$ with the adjacency matrix $\mathbf{A}$ as an example for the topology modification. It is straight-forward to generalize that to the feature modification with feature matrix $\mathbf{X}$.

The discretized modification is guided by a hyper-gradient-based score matrix:

\vspace{-4mm}
\begin{equation}
    \mathbf{S} = (-\nabla_{\mathbf{A}})\circ(\mathbf{1}-2\mathbf{A}),
    \label{eq:score matrix}
\end{equation}
\vspace{-4mm}

\noindent where $\circ$ denotes Hadamard product, $\mathbf{1}$ is an all-one matrix. This score matrix is composed by `preference' (i.e., $-\nabla_{\mathbf{A}}$) and `modifiability' (i.e., $(\mathbf{1}-2\mathbf{A})$). Only entries with both high `preference' and `modifiability' can be assigned with high scores. For example, large positive $(-\nabla_{\mathbf{A}})[i,j]$ indicates strong preference of adding an edge between the $i$-th and the $j$-th nodes based on the hyper-gradient and if there was no edge between the $i$-th and the $j$-th nodes (i.e., $\mathbf{A}[i,j]=0$), the signs of $(-\nabla_{\mathbf{A}})[i,j]$ and $(\mathbf{1}-2\mathbf{A})[i,j]$ are the same which result in a large $\mathbf{S}[i,j]$. Then, corresponding $B$ entries in $\mathbf{A}$ are modified by flipping the sign of them based on the indices of the top-$B$ entries in $\mathbf{S}$.



The continuous modification is hyper-gradient descent with budget-adaptive learning rate:

\vspace{-3mm}
\begin{equation}
    \mathbf{A} \leftarrow \mathbf{A}-\frac{B}{\sum_{i,j}|\nabla_{\mathbf{A}}|[i,j]} \cdot \nabla_{\mathbf{A}}.
    \label{eq:continuous updating}
\end{equation}
\vspace{-2mm}

\noindent We compute the learning rate based on the ratio of the modification budget $B$ to the sum of absolute values of the hyper-gradient matrix. In implementation, for both modification methods, we set the budget in every iteration as $b$ and update the graph in multiple steps until we run out of the total budget $B$ so as to balance the effectiveness and the efficiency. Algorithm~\ref{alg:sanitation procedure} summarizes the detailed modification procedure. In addition, in our experiments, the $B$ for topology ($B_{\texttt{topo}}$) and the $B$ for feature ($B_{\texttt{fea}}$) are set separately since the modification cost on different elements of a graph may not be comparable.

{\em Remarks.} We notice that LDS~\cite{franceschi2019learning} formulates the graph learning problem for GNN under the bilevel optimization context. Here, we claim the differences and advantages: (1) LDS focuses on learning graph topology, but \gasoline~can handle any graph components (e.g., topology and feature); (2) LDS formulates the topology as a set of Bernoulli random variables, whose updating requires multiple samplings which are time-consuming, but \gasoline~works in a deterministic and efficient way and also provides the discrete solutions; (3) in the following section we will introduce a speed-up variant of \gasoline~which shows great efficacy.



\begin{algorithm}
\caption{\gasoline}
\label{alg:sanitation procedure}
\SetAlgoLined
\SetKwInOut{Input}{Input}\SetKwInOut{Output}{Output}
\Input{the given graph $G$, the labeled nodes $\mathcal{Z}$, modification budget $B$, budget $b$ in every modification step, number of training/validation split fold $K$, truncating iteration $P$ and converging iteration $T$;}
\Output{the modified graph $\tilde{G}$;}

initialization: split the labeled nodes $\mathcal{Z}$ and their labels into $K$ folds: $\mathcal{Z}=\{\mathcal{Z}^{1},\dots,\mathcal{Z}^{K}\}$, $\mathcal{Y}_{\texttt{labeled}}=\{\mathcal{Y}^{1},\dots,\mathcal{Y}^{K}\}$; modified graph $\tilde{G}\leftarrow G$; cumulative budget $\delta\leftarrow 0$\;

\While {$\delta<B$}{
\For {k=1 to K}
{$\mathcal{V} \leftarrow \mathcal{Z}^{k}$, $\mathcal{T} \leftarrow \mathcal{Z}\backslash \mathcal{Z}^{k}$, $\mathcal{Y}_{\texttt{valid}} \leftarrow \mathcal{Y}^{k}$, $\mathcal{Y}_{\texttt{train}} \leftarrow \mathcal{Y}_{\texttt{labeled}}\backslash \mathcal{Y}^{k}$, $\nabla_{\tilde{G}}^k\mathcal{L} \leftarrow 0$\;
\For{$t=1$ to $T$}{
update $\theta^{t}$ to $\theta^{t+1}$ by Eq.~\eqref{eq:update theta}\;
\If{$t>P$}{
compute $\nabla_{\tilde{G}}^k\mathcal{L}_{\texttt{valid}}$ given $\{\tilde{G}, \theta^{t+1},\mathcal{V},\mathcal{Y}_{\texttt{valid}}\}$\;
$\nabla_{\tilde{G}}^k\mathcal{L} \leftarrow \nabla_{\tilde{G}}^k\mathcal{L}+\nabla_{\tilde{G}}^k\mathcal{L}_{\texttt{valid}}$
}
}
}
calibrate hyper-gradients $\{\nabla_{\tilde{\mathbf{A}}}^k\mathcal{L}\}$ by Eq.~\eqref{eq:fix derivative w.r.t. A} (if needed)\;
sum hyper-gradients $\{\nabla_{\tilde{G}}^k\mathcal{L}\}$ into $\nabla_{\tilde{G}}$\;
update $\tilde{G}$ based on the guide of score matrix $\mathbf{S}$ by Eq.~\eqref{eq:score matrix} (discretized modification) or by Eq.~\eqref{eq:continuous updating} (continuous modification) with budget $b$\;
$\delta\leftarrow\delta+b$
}
\end{algorithm}

\noindent \textbf{D - Speedup and Scale-up.} 
The core operation of our proposed \gasoline~is to compute hyper-gradient w.r.t. the graph components (i.e., $\mathbf{A}$ and $\mathbf{X}$) which leads into a gradient matrix (e.g. $\nabla_{\mathbf{A}}\mathcal{L}$).
In many real-world scenarios (e.g., malfunctions of certain nodes, targeted adversarial attacks), perturbations are often around a small set of nodes, which leads to low-rank perturbation matrices. Hence, for topology modification, we propose to further decompose the incremental matrix (i.e., $\Delta \mathbf{A}$) into its low-rank representation (i.e., $\Delta \mathbf{A} = \mathbf{U}\mathbf{V}^{\prime}$), and compute the hyper-gradient with respect to the low-rank matrices instead, which can significantly speedup and scale up the computation. Recall that the low-rank assumption is only held for the incremental matrix, but for the modified graph (i.e., $\tilde{\mathbf{A}}$), it is not limited to be low-rank. Mathematically, the low-rank modification can be represented as:
 



\vspace{-3mm}
\begin{equation}
    \tilde{\mathbf{A}} = \mathbf{A}+\Delta \mathbf{A} = \mathbf{A}+\mathbf{U}\mathbf{V}^{\prime},
\end{equation}
\vspace{-4mm}



\noindent where $\mathbf{U}$, $\mathbf{V}\in\mathbb{R}^{n\times r}$, and $r$ is the rank of $\Delta \mathbf{A}$. Hence, by substituting $\mathbf{A}$ with $\mathbf{A}+\mathbf{U}\mathbf{V}^{\prime}$ in Eq.~\eqref{eq:graph sanitation for semi-supervised node classification} (i.e., $G=\{\mathbf{A}+\mathbf{U}\mathbf{V}^{\prime}, \mathbf{X}\}$) and changing the optimization variable from $\mathbf{A}$ into $\mathbf{U}$ and $\mathbf{V}$, we can obtain hyper-gradient with respect to $\mathbf{U}$ and $\mathbf{V}$ (i.e., $\nabla_{\mathbf{U}}\mathcal{L}$ and $\nabla_{\mathbf{V}}\mathcal{L}$) in the same manner as Eq.\eqref{eq:hyper-gradient approximation}. By aggregating the hyper-gradients from different training/validation splits as we introduced in Sec.~\ref{sec:method}-B, we obtain aggregated hyper-gradients $\nabla\mathbf{U}$ and $\nabla\mathbf{V}$. Any gradient descent-based method can then be used to update $\mathbf{U}$ and $\mathbf{V}$.  

In this way, we can significantly reduce the time and space complexity, which is summarized in the following lemma. Notice that $n,m,d$ are number of nodes, number of edges and feature dimension, respectively and we have $d\ll n$ and $m \ll n^2$. As a comparison, the time complexity of computing $\nabla_{\mathbf{A}}\mathcal{L}$ is $O(n^2d)$ and the space complexity of computing $\nabla_{\mathbf{A}}\mathcal{L}$ is $O(n^2)$. Hence, this low-rank method is much more efficient in both time and space.

\vspace{-2mm}
\begin{lemma}
\label{lm:time and space complexity} For computing $\nabla_{\mathbf{U}}\mathcal{L}$ and $\nabla_{\mathbf{V}}\mathcal{L}$, the time complexity is $O(nd^2+md)$ and the space complexity is $O(m+nd)$. 
\end{lemma}

\vspace{-4mm}
\begin{proof}
See Appendix.
\end{proof}





\section{Experiments}
\label{sec:experiments}

In this section, we perform empirical evaluations. All the experiments are designed to answer  the following research questions:
\begin{itemize}
    \item [$\bullet$\ \textbf{RQ1}] How applicable is the proposed \gasoline\ with respect to different backbone/downstream classifiers, as well as different modification strategies? 
    \item [$\bullet$\ \textbf{RQ2}] How effective is the proposed \gasoline\ for initial graphs under various forms of perturbation? To what extent does the proposed \gasoline\ strengthen the existing robust GNNs?
    \item [$\bullet$\ \textbf{RQ3}] How efficient and effective is the low-rank \gasoline?
\end{itemize}

\vspace{-1mm}
\subsection{Experiment Setups}
\label{sec:experiment setups}
We evaluate the proposed \gasoline~ on Cora, Citeseer, and Polbolgs datasets~\cite{kipf2016semi,zugner2018adversarial,DBLP:conf/iclr/ZugnerG19}. Since the Polblogs dataset does not contain node features, we use an identity matrix as the node feature matrix. All the datasets are undirected unweighted graphs and we experiment on the largest connected component of every dataset.

In order to set fair modification budgets across different datasets, the modification budget on adjacency matrix $B_{\texttt{topo}}$ is defined as $B_{\texttt{topo}}=m\times\texttt{modification rate}_{\texttt{topo}}$ and the budget on feature matrix $B_{\texttt{fea}}$ is defined as $B_{\texttt{fea}}=n\times d\times\texttt{modification rate}_{\texttt{fea}}$.

\noindent where $m$ is the number of existing edges; $n$ is the number nodes; $d$ is the node feature dimension. We set $\texttt{modification rate}_{\texttt{topo}}=0.1$ and $\texttt{modification rate}_{\texttt{fea}}=0.001$ throughout all the experiments. Detailed hyper-parameter settings are attached in Appendix.


We use the accuracy as the evaluation metric and repeat every set of experiment $10$ times to report the mean $\pm$ std value.


\subsection{Applicability of \gasoline}
In this subsection, we conduct an in-depth study about the property of modified graphs by \gasoline. The proposed \gasoline\ trains a \emph{backbone classifier} in the lower-level problem and uses the trained backbone classifier to modify the initially-provided graph and improve the performance of the \emph{downstream classifier} on the test nodes. In addition, \gasoline\ is capable of modifying both the graph topology (i.e., $\mathbf{A}$) and feature (i.e., $\mathbf{X}$) in both the discretized and continuous fashion. To verify that, we select three classic GNNs-based node classifiers, including GCN~\cite{kipf2016semi}, SGC~\cite{wu2019simplifying}, and APPNP~\cite{klicpera2018predict} to serve as the backbone classifiers and the downstream classifiers. The detailed experiment procedure is as follows. First, we modify the given graph using proposed \gasoline~ algorithm with $4$ modification strategies (i.e., modifying topology or node feature with discretized or continuous modification). Each variant is implemented with $3$ backbone classifiers so that in total there are $12$ sets of \gasoline\ settings.  Second, with the $12$
modified graphs, we test the performance of $3$ downstream classifiers and report the result (mean$\pm$std Acc) under each setting. For the results in this subsection, the initially provided graph is Citeseer~\cite{kipf2016semi}.

Experimental results are reported in Table~\ref{tab:transferability} where `DT' denotes `discretized topology modification', `CT' denotes `continuous topology modification', `DF' denotes `discretized feature modification', and `CF' denotes `continuous feature modification'. The second row of Table~\ref{tab:transferability} shows the results on the initially-provided graph and the other rows denote the results on modified graphs with different settings. We use $\bullet$ to indicate that the improvement of the result is statistically significant compared with results on the initially-provided graph with a $p$-value$<0.01$, and we use $\circ$ to indicate no statistically significant improvement. We have the following observations. First, in the vast majority cases, the proposed \gasoline\ is able to statistically significantly improve the accuracy of the downstream classifier over the initially-provide graph, for every combination of the modification strategy (discretized vs. continuous) and the modification target (topology vs. feature). 
Second, 
the graphs modified by \gasoline\ with different backbone classifier can benefit different downstream classifiers, which demonstrates great transferability and broad applicability.


\begin{table}[t!]
\resizebox{0.85\linewidth}{!}{
\begin{tabular}{cc|ccc}
\toprule
\textbf{Variant}          & \textbf{Backbone} & \textbf{GCN} & \textbf{SGC} & \textbf{APPNP} \\ \midrule
None                      & None              & 72.2$\pm$0.5   & 72.8$\pm$0.2   & 71.8$\pm$0.4     \\ \midrule
\multirow{3}{*}{DT} & GCN               & 74.7$\pm$0.3$\bullet$   & 74.8$\pm$0.1$\bullet$   & 75.4$\pm$0.2$\bullet$     \\
                          & SGC               & 74.7$\pm$0.4$\bullet$   & 75.2$\pm$0.2$\bullet$   & 75.6$\pm$0.3$\bullet$     \\
                          & APPNP             & 74.6$\pm$0.3$\bullet$   & 74.6$\pm$0.1$\bullet$   & 75.4$\pm$0.4$\bullet$     \\ \midrule
\multirow{3}{*}{DF}  & GCN               & 72.4$\pm$0.3$\circ$   & 72.7$\pm$0.2$\circ$   & 72.8$\pm$0.4$\bullet$     \\
                          & SGC               & 73.3$\pm$0.5$\bullet$   & 73.4$\pm$0.2$\bullet$   & 73.6$\pm$0.4$\bullet$     \\
                          & APPNP             & 72.6$\pm$0.3$\circ$   & 72.9$\pm$0.1$\circ$   & 73.6$\pm$0.4$\bullet$     \\ \midrule
\multirow{3}{*}{CT} & GCN               & 73.1$\pm$0.4$\bullet$   & 73.6$\pm$0.1$\bullet$   & 74.8$\pm$0.2$\bullet$     \\
                          & SGC               & 73.0$\pm$0.3$\bullet$   & 73.5$\pm$0.2$\bullet$   & 74.4$\pm$0.3$\bullet$     \\
                          & APPNP             & 72.8$\pm$0.5$\circ$   & 73.4$\pm$0.1$\bullet$   & 74.4$\pm$0.9$\bullet$     \\ \midrule
\multirow{3}{*}{CF}  & GCN               & 72.7$\pm$0.4$\circ$   & 73.6$\pm$0.1$\bullet$   & 73.8$\pm$0.3$\bullet$     \\
                          & SGC               & 72.9$\pm$0.4$\bullet$   & 73.6$\pm$0.4$\bullet$   & 73.8$\pm$0.4$\bullet$     \\
                          & APPNP             & 73.0$\pm$0.3$\bullet$   & 73.6$\pm$0.2$\bullet$   & 73.9$\pm$0.3$\bullet$     \\ \bottomrule
\end{tabular}
}
\caption{Effectiveness of \gasoline~ under multiple variants (Mean$\pm$Std Accuracy). The first and second columns denote the modification strategies and backbone classifiers adopted by \gasoline~ respectively. The remaining columns show the performance of various downstream classifiers. $\bullet$ indicates significant improvement compared with results on the original graph (values at the second row) with a $p$-value<0.01 and $\circ$ indicates no statistically significant improvement.}
\label{tab:transferability}
\end{table}

\begin{figure}[t!]
\begin{subfigure}{.23\textwidth}
  \centering
  \includegraphics[width=\linewidth]{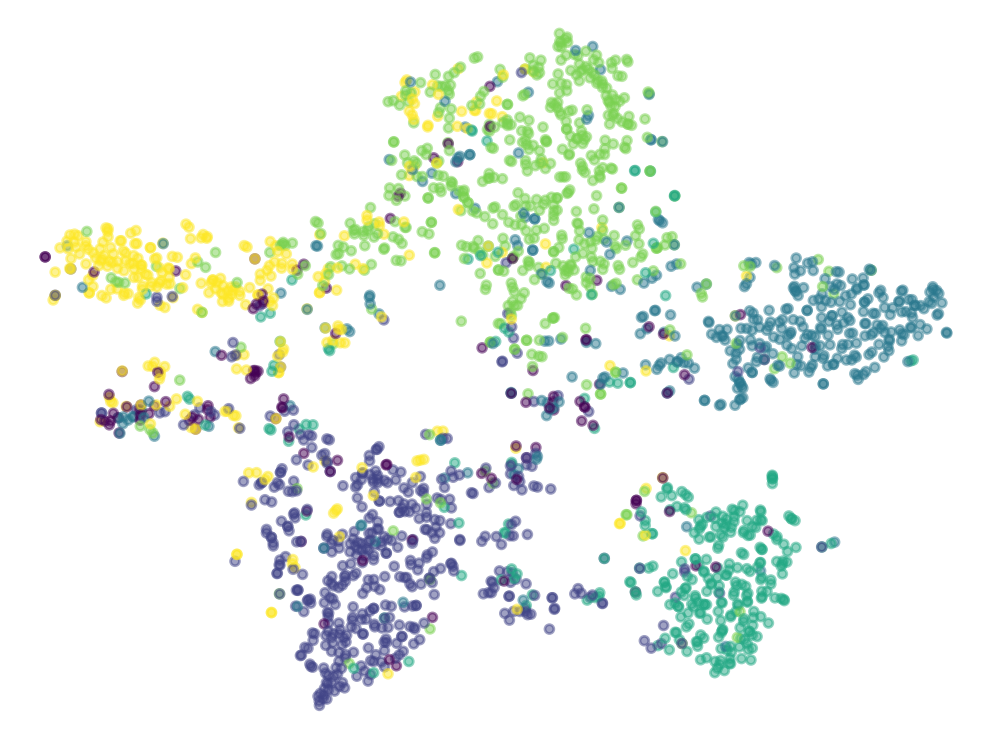}
  \vspace{-4mm}
  \caption{Original}
  \label{fig:sub-original}
\end{subfigure}
\begin{subfigure}{.23\textwidth}
  \centering
  \includegraphics[width=\linewidth]{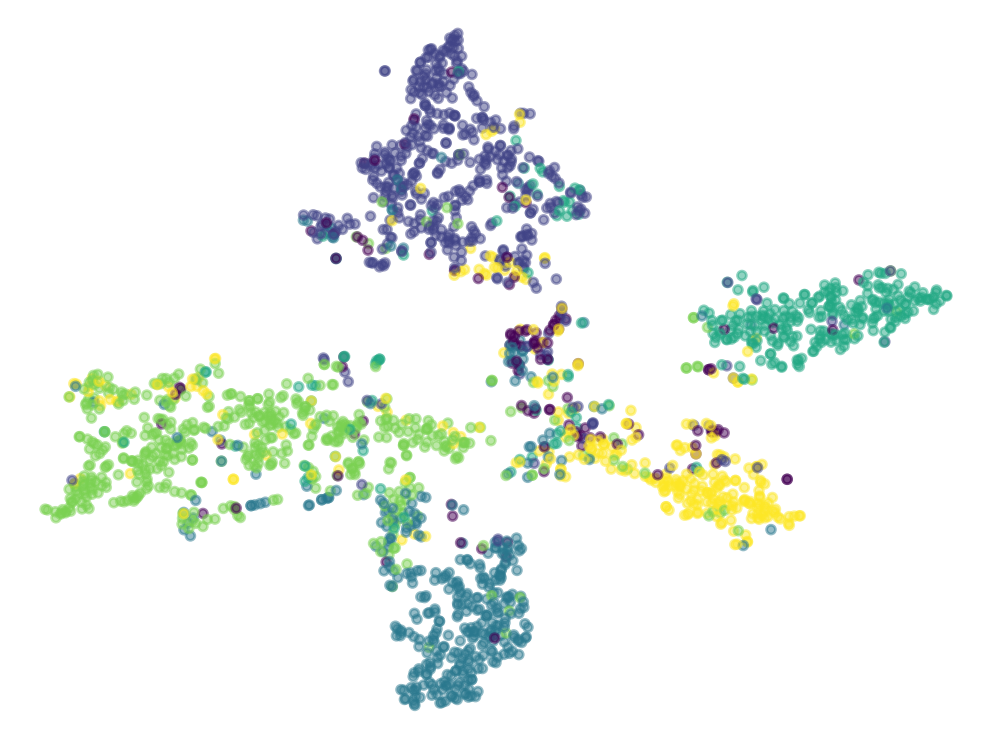}  
  \vspace{-4mm}
  \caption{After modification}
  \label{fig:sub-SGC}
\end{subfigure}
\vspace{-4mm}
\caption{Visualization of node embeddings from original Citeseer graph (a) and modified Citeseer graph by \gasoline~ (b). Best viewed in color.}
\label{fig:visualization results}
\end{figure}

We further provide visualization of node embeddings before and after modification. We present the visualizations of initial Citeseer graph and the modified Citeseer graph from \gasoline~ discretized topology modification variant with backbone classifier as SGC~\cite{wu2019simplifying}. The detailed visualization procedure is that we utilize the training set $\mathcal{T}$ (and corresponding labels $\mathcal{Y}_{\texttt{train}}$) of given initial/modified graphs to train a GCN~\cite{kipf2016semi} and use hidden representation of the trained GCN to encode every node into a high-dimensional vector. Then, we adopt t-SNE~\cite{van2014accelerating} method to map the high-dimensional node embeddings into two-dimensional ones for visualization. Figure~\ref{fig:visualization results} shows the visualization results of node embeddings of the original Citeseer graph and the modified Citeseer graph. Clearly, the node embeddings from modified graph are more discriminative than the embeddings from the original graph. In specific, the clusters are more cohesive and there is less overlap between clusters in the modified graphs (i.e., Figures~\ref{fig:sub-SGC}) compared with those on the original graph (i.e., Figure~\ref{fig:sub-original}). It further demonstrates that even we have no knowledge about the downstream classifiers (in this case the backbone classifier and downstream classifier are different), the proposed \gasoline~ can still improve the graph quality to benefit downstream classifiers.

\begin{figure*}[t!]
\begin{subfigure}{.97\textwidth}
  \centering
  \includegraphics[width=.99\linewidth]{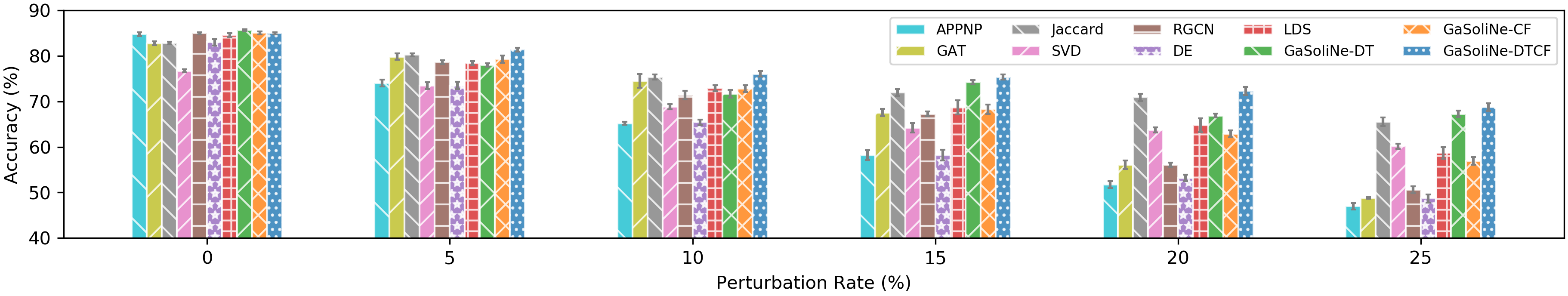}
  \vspace{-2mm}
  \caption{metattack}
  \label{fig:meta attack bar}
\end{subfigure}
\begin{subfigure}{.97\textwidth}
  \centering
  \includegraphics[width=.99\linewidth]{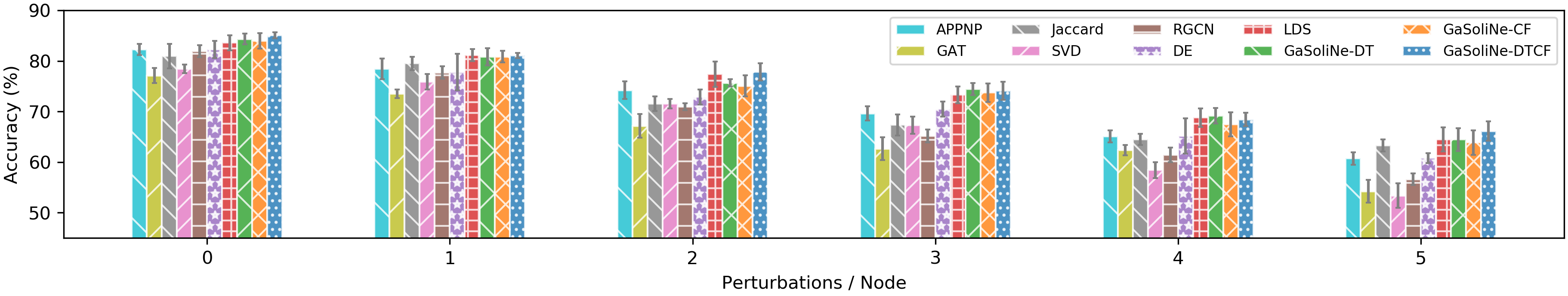}  
  \vspace{-2mm}
  \caption{\nettack}
  \label{fig:nettack bar}
\end{subfigure}
\begin{subfigure}{.97\textwidth}
  \centering
  \includegraphics[width=.99\linewidth]{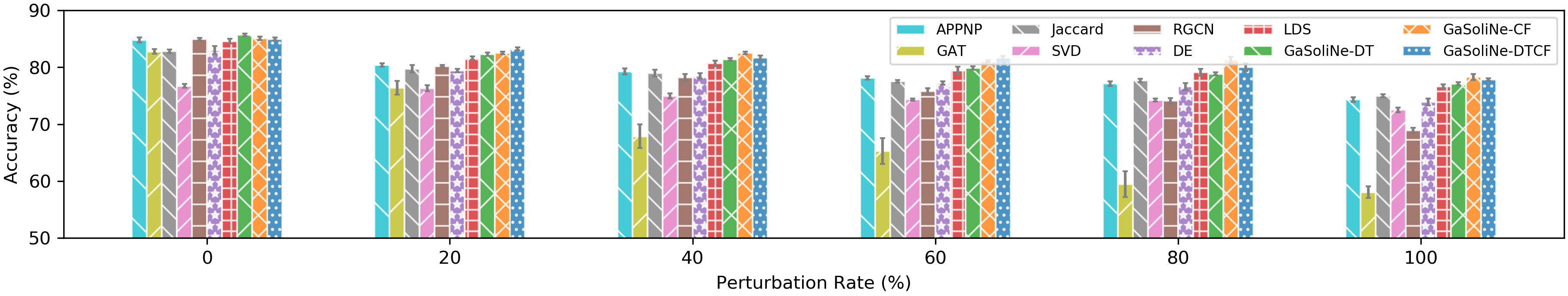}  
  \vspace{-2mm}
  \caption{random attack}
  \label{fig:random attack bar}
\end{subfigure}
\vspace{-4mm}
\caption{Performance of models under (a) metattack, (b) \nettack, and (c) random attack. Best viewed in color.}
\vspace{-2mm}
\label{fig:baseline on one dataset}
\end{figure*}

\begin{table*}[t!]
\resizebox{0.9\linewidth}{!}{
\begin{tabular}{cc|cccccccccc}
\toprule
\textbf{Attack}                & \textbf{Data} & \textbf{APPNP} & \textbf{GAT}      & \textbf{Jaccard} & \textbf{SVD}      & \textbf{RGCN} & \textbf{DE} & \textbf{LDS} & \textbf{G-DT}     & \textbf{G-CF}     & \textbf{G-DTCF}   \\ \midrule
\multirow{3}{*}{metattack}     & Cora          & 47.0$\pm$0.7       & 48.8$\pm$0.2          & 65.4$\pm$0.9         & 60.3$\pm$0.8          & 50.6$\pm$0.8      & 48.7$\pm$0.9    & 58.7$\pm$1.3     & 67.3$\pm$0.7          & 57.0$\pm$0.9          & \textbf{68.8$\pm$0.9} \\
                               & Citeseer      & 49.4$\pm$2.2       & 62.4$\pm$0.7          & 57.1$\pm$1.0         & 49.5$\pm$0.8          & 55.5$\pm$1.4      & 50.1$\pm$2.3    & 58.2$\pm$2.3     & \textbf{63.5$\pm$1.5} & 58.4$\pm$1.5          & 62.2$\pm$1.0          \\
                               & Polblogs      & 58.4$\pm$3.6       & 48.2$\pm$6.6          & N/A              & \textbf{79.1$\pm$2.4} & 50.8$\pm$0.9      & 56.4$\pm$6.3    & 63.7$\pm$5.7     & 65.0$\pm$0.7          & 55.0$\pm$4.1          & 64.7$\pm$1.4          \\ \midrule
\multirow{3}{*}{Nettack}       & Cora          & 60.7$\pm$1.2       & 54.2$\pm$2.3          & 63.7$\pm$1.4         & 52.9$\pm$2.8          & 56.5$\pm$1.1      & 60.8$\pm$1.0    & 64.5$\pm$2.4     & 64.5$\pm$2.2          & 63.9$\pm$2.4          & \textbf{66.1$\pm$1.9} \\
                               & Citeseer      & 68.3$\pm$6.8       & 61.9$\pm$4.4          & 72.5$\pm$3.3         & 50.2$\pm$6.6          & 56.4$\pm$1.5      & 63.3$\pm$4.7    & 71.0$\pm$3.3     & 71.6$\pm$3.9          & 69.4$\pm$4.8          & \textbf{74.3$\pm$1.6} \\
                               & Polblogs      & 90.5$\pm$1.0       & 91.1$\pm$0.7          & N/A              & \textbf{93.6$\pm$1.2} & 93.1$\pm$0.2      & 89.1$\pm$2.4    & 91.1$\pm$1.8     & 92.3$\pm$1.6          & 90.3$\pm$0.7          & 92.4$\pm$1.7          \\ \midrule
\multirow{3}{*}{\makecell{random\\attack}} & Cora          & 74.3$\pm$0.4       & 58.1$\pm$1.0          & 75.1$\pm$0.5         & 72.6$\pm$0.3          & 68.9$\pm$0.4      & 73.9$\pm$0.6    & 76.6$\pm$0.4     & 77.1$\pm$0.3          & \textbf{78.3$\pm$0.5} & 77.8$\pm$0.2          \\
                               & Citeseer      & 69.8$\pm$0.6       & 60.8$\pm$1.6          & 69.7$\pm$0.5         & 66.7$\pm$0.4          & 65.7$\pm$0.2      & 69.4$\pm$0.5    & 72.3$\pm$0.4     & \textbf{73.8$\pm$0.2} & 72.3$\pm$0.4          & 73.4$\pm$0.5          \\
                               & Polblogs      & 74.7$\pm$2.8       & \textbf{84.5$\pm$1.0} & N/A              & 83.3$\pm$2.8          & 81.7$\pm$0.9      & 75.9$\pm$1.4    & 73.2$\pm$2.8     & 73.4$\pm$4.1          & 77.1$\pm$1.6          & 77.6$\pm$2.9         
\\ \bottomrule
\end{tabular}
}
\caption{Comparison with baselines on heavily poisoned datasets (Mean$\pm$Std Accuracy). Some results are not applicable since Jaccard requires node features which are absent on Polblogs graph. G denotes \gasoline~for short.}
\vspace{-7mm}
\label{tab:with baselines on heavily poisoned datasets}
\end{table*}

\vspace{-2mm}
\subsection{Effectiveness of \gasoline}
\label{sec:effectivenss of gasoline}


As we point out in Sec.~\ref{sec:introduction}, the defects of the initially-provided graph could be due to various reasons. 
In this subsection, we evaluate the effectiveness of the proposed \gasoline~ by (A) the comparison with baseline methods on various poisoned/noisy graphs and (B) integrating with existing robust GNNs methods. The attack methods we use to poison benign graphs are as follows: (1) Random Attack randomly flips entries of benign adjacency matrices with different $\texttt{perturbation rate}$; (2) \nettack~\cite{zugner2018adversarial} attacks a set of target nodes with different \texttt{perturbations/node}; (3) metattack~\cite{DBLP:conf/iclr/ZugnerG19} poisons the performance of node classifiers by perturbing the overall benign graph topology with different $\texttt{perturbation rate}$. 
    
    


\noindent \textbf{A - Comparison with baseline methods.}
We compare \gasoline~ with the following baseline methods: APPNP~\cite{klicpera2018predict}, GAT~\cite{velivckovic2018graph}, Jaccard~\cite{wu2019adversarial}, SVD~\cite{entezari2020all}, RGCN~\cite{zhu2019robust}, DE~\cite{rong2019dropedge}, and LDS~\cite{franceschi2019learning}. Recall that we feed all the graph modification-based methods (Jaccard, SVD, DE, LDS, \gasoline) with the exactly same downstream classifier (APPNP) for a fair comparison.

We set $3$ variants of \gasoline~ to compare with the above baselines. To be specific, we refer to (1) \gasoline\ with discretized modification on topology as \gasoline-DT, (2) \gasoline\ with continuous modification on feature as \gasoline-CF, and (3) \gasoline\ with discretized modification on topology and continuous modification on feature as \gasoline-DTCF. All these \gasoline~ variants use APPNP~\cite{klicpera2018predict} as both the backbone classifier and the downstream classifier. We test various perturbation rates (i.e., \texttt{perturbation rate} of metattack from $5\%$ to $25\%$ with a step of $5\%$, \texttt{perturbation rate} of random attack from $20\%$ to $100\%$ with a step of $20\%$, and \texttt{perturbations/node} of \nettack~ from $1$ to $5$) to attack the Cora~\cite{kipf2016semi} dataset and report the accuracy (mean$\pm$std) in Figure~\ref{fig:baseline on one dataset}. From experiment results we observe that: (1) with the increase of adversarial perturbation, the performance of all methods drops, which is consistent with our intuition; (2) variants of \gasoline~ consistently outperform the baselines under various adversarial/noisy scenarios; and (3) the proposed \gasoline\ even improves over the original, benign graphs (i.e., $0$ $\texttt{perturbation rate}$ and $0$ $\texttt{perturbations/node}$).

An interesting question is, if the initially-provided graph is heavily poisoned/noisy, to what extent is the proposed \gasoline\ still effective? To answer this question, we study the performance of \gasoline~ and other baseline methods on heavily-poisoned graphs ($100\%$ \texttt{perturbation rate} of random attack,  $25\%$ \texttt{perturbation rate} of metattack, and $5$ \texttt{perturbations/node} of \nettack). The detailed experiment results are presented in Table~\ref{tab:with baselines on heavily poisoned datasets}. In most cases, \gasoline~ can obtain competitive or even better performance against baseline methods.  On the Polblogs graph, \gasoline\ does not perform as well as in the other two datasets. This is because, (1) the Polblogs graph does not have node feature which weakens the effectiveness of modification from \gasoline~and (2) the Polblogs graph has strong low-rank structure, which can be further verified in Sec.~\ref{sec:low-rank exp}. As flexible solutions, in the next subsection, we study that if \gasoline~ can work together with other graph defense methods.

\noindent \textbf{B - Incorporating with graph defense strategies.}
\gasoline~ does not make any assumption about the property of the defects of the initially-provided graph. 
We further evaluate if \gasoline~ can help boost the performance of both model-based and data-based defense strategies under the heavily-poisoned settings. We use a data-based defense baseline SVD~\cite{entezari2020all}, a model-based defense baseline RGCN~\cite{zhu2019robust}, and another strong baseline GAT~\cite{velivckovic2018graph} to integrate with \gasoline~ since they have shown competitive performance from Table~\ref{tab:with baselines on heavily poisoned datasets} and Figure~\ref{fig:baseline on one dataset}. The detailed procedure is that for model-based methods (i.e., GAT and RGCN), \gasoline~ modifies the graph at first, and then the baselines are implemented on the modified graphs to report the final results. For the data-based method (i.e., SVD), we first implement the baseline to preprocess graphs, and then we modify graphs again by \gasoline, and finally run the downstream classifiers (APPNP) on the twice-modified graphs. In this task, specifically, we use \gasoline-DTCF to integrate various defense methods. 
In order to heavily poison the graphs, we use metattack~\cite{DBLP:conf/iclr/ZugnerG19} with $\texttt{perturbation rate}$ as $25\%$ to attack the benign graphs. We report the results in Table~\ref{tab:improve existing methods} and observe that after integrating with \gasoline, performance of all the defense methods further improves significantly with a $p$-value<0.01.

{\em Remark.} A detailed case study about the behaviour of \gasoline~ and a set of sensitivity studies for the hyper-parameters are provided in Appendix.

\subsection{Efficacy of Low-Rank~\gasoline}
\label{sec:low-rank exp}

\begin{wrapfigure}{R}{0.26\textwidth}
\centering
\includegraphics[width=0.26\textwidth]{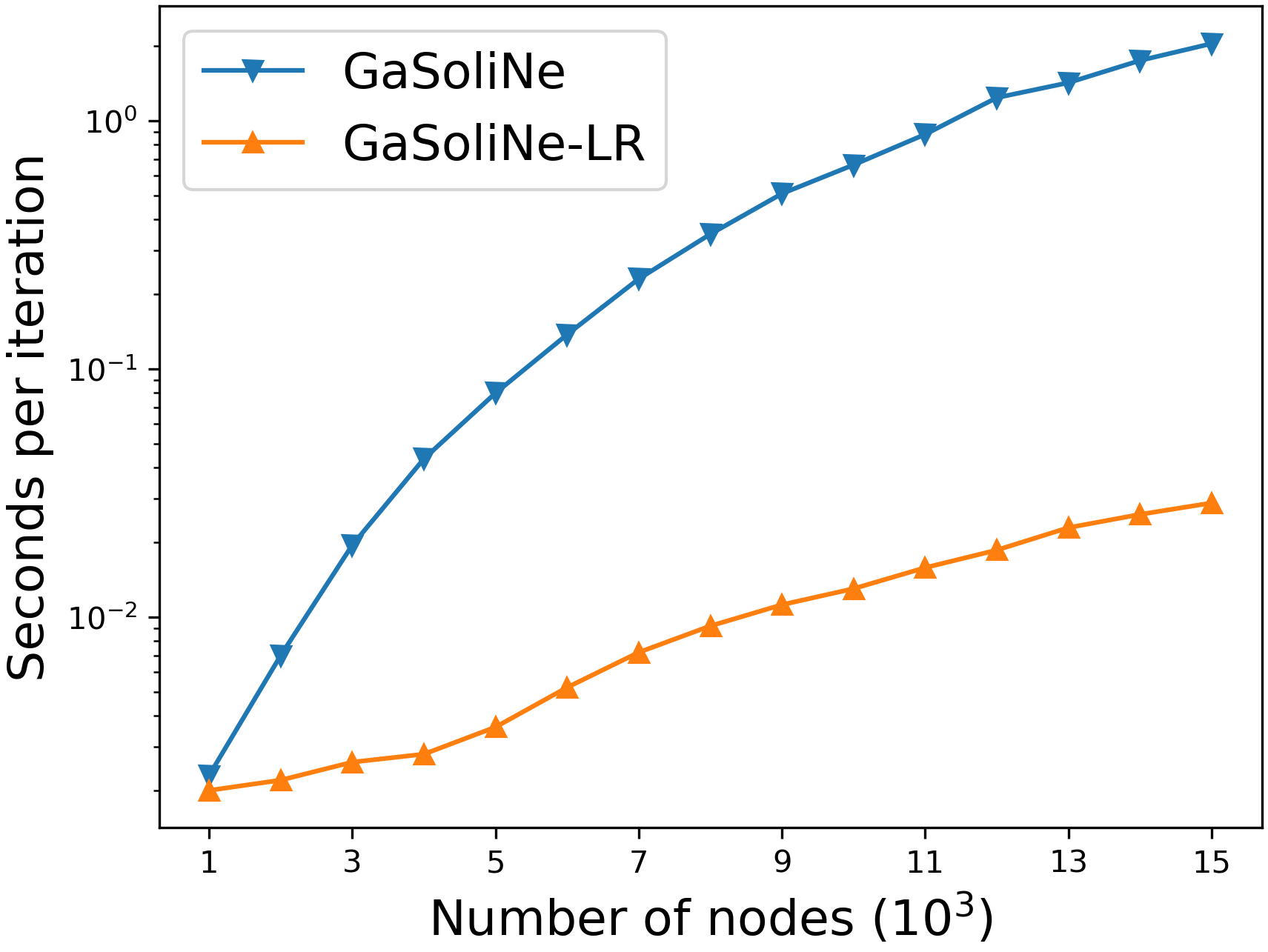}
\caption{Efficiency comparison between \gasoline~and \gasoline-LR} \label{fig:efficiency study}
\end{wrapfigure}
In order to answer RQ3, we first compare the performance of APPNP~\cite{klicpera2018predict} on two modified graphs from the low-rank \gasoline~(denoted as \gasoline-LR) and the original \gasoline, respectively. Specifically, for the original \gasoline, we adopt its variant with continuous modification towards the network topology. Due to the space limitation, we only show the results given graphs perturbed by metattack~\cite{DBLP:conf/iclr/ZugnerG19} in Table~\ref{tab:performance comparison between original and low-rank gasoline}. We observe that in most settings on both the Cora and Citeseer datasets, the \gasoline-LR can obtain promising performance against the original \gasoline. Surprisingly, on the Polblogs dataset, the \gasoline-LR shows great advantages over the original \gasoline. One possible explanation is that Polblogs dataset is inherently low-rank (which can be corroborated by Table~\ref{tab:with baselines on heavily poisoned datasets} where SVD~\cite{entezari2020all} obtains strong performance) and \gasoline-LR learns a low-rank incremental matrix which amplifies the advantage further.

To verify the efficiency of proposed \gasoline-LR, we generate a set of synthetic graphs with different number of nodes $n$. The wall-clock time for computing hyper-gradient is presented in Figure~\ref{fig:efficiency study}. Clearly, the \gasoline-LR is much more efficient compared with the original \gasoline~especially when the network size is large.

\vspace{-1.5mm}
\section{Related Work}
\label{sec:related work}

\noindent\textbf{A - Graph Modification}
The vast majority of the existing works on graph modification assume the initially provided graph is impaired or perturbed in a specific way. %
Network imputation problems focus on restoring missing links in a partially observed graph. 
For example, Liben-Nowell and Kleinberg~\cite{liben2007link} study the property of a set of node topology proximity measures; Huisman~\cite{huisman2009imputation} handles missing data in exponential random graphs. Besides, knowledge graph completion is to predict missing links between entities. The representative works include TransE~\cite{bordes2013translating}, TransH~\cite{wang2014knowledge}, ComplEx~\cite{trouillon2016complex} and many more. In another line of research, network enhancement and denoising problems delete irrelevant links for the given graphs. Arts such as NE~\cite{wang2018network}, E-net~\cite{xu2020robust}, Mask-GVAE~\cite{li2021mask} study this problem under various scenarios. For network connectivity analysis, Chen et al.~\cite{chen2015node,chen2018network} modify the underlying topology in order to manipulate the graph connectivity. 
Another relevant line is adversarial defense, which is as a response for the attack on graph mining models. Wu et al.~\cite{wu2019adversarial} claim that deleting edges connecting two dissimilar nodes is an effective defense strategy; Entezari et al.~\cite{entezari2020all} utilize the low-rank approximation of given graphs to retain the performance of downstream GCNs; Jin et al.~\cite{DBLP:conf/kdd/Jin0LTWT20} merge the topology sparsity and feature smoothness into the optimization goal and jointly learn the downstream classifiers. In addition, supervised PageRank~\cite{backstrom2011supervised,li2016quint} and constrained spectral clustering~\cite{wang2010flexible} also encode extra supervision to guide the modification of graphs.

\begin{table}[t!]
\resizebox{0.9\linewidth}{!}{
\begin{tabular}{ccccc}
\toprule
\textbf{Data}             & \textbf{Ptb Rate (\%)} & \textbf{APPNP} & \textbf{\gasoline} & \textbf{\gasoline-LR} \\
\midrule
\multirow{6}{*}{Cora}     & 0                      & 84.0$\pm$0.4       & 85.2$\pm$0.2           & 84.4$\pm$0.3          \\
                          & 5                      & 74.1$\pm$0.7       & 77.4$\pm$0.5           & 75.0$\pm$0.3          \\
                          & 10                     & 65.2$\pm$0.4       & 70.8$\pm$0.5           & 67.9$\pm$0.9          \\
                          & 15                     & 58.2$\pm$1.1       & 67.1$\pm$0.8           & 65.3$\pm$0.8          \\
                          & 20                     & 51.7$\pm$0.7       & 62.5$\pm$0.5          & 60.2$\pm$1.2          \\
                          & 25                     & 47.0$\pm$0.7       & 57.3$\pm$0.6           & 57.1$\pm$0.5          \\ \midrule
\multirow{6}{*}{Citeseer} & 0                      & 71.8$\pm$0.4       & 74.7$\pm$0.2           & 73.4$\pm$0.2          \\
                          & 5                      & 67.6$\pm$0.9       & 69.6$\pm$0.7           & 68.2$\pm$0.8          \\
                          & 10                     & 61.8$\pm$0.8       & 66.3$\pm$1.0           & 63.9$\pm$0.4          \\
                          & 15                     & 54.1$\pm$0.8       & 59.3$\pm$1.1           & 56.8$\pm$1.1          \\
                          & 20                     & 51.0$\pm$1.2       & 56.5$\pm$0.9           & 55.3$\pm$0.9          \\
                          & 25                     & 49.4$\pm$2.2       & 57.7$\pm$1.8           & 56.5$\pm$0.8          \\ \midrule
\multirow{6}{*}{Polblogs} & 0                      & 94.1$\pm$0.6       & 95.3$\pm$0.6           & 95.7$\pm$0.3          \\
                          & 5                      & 70.1$\pm$0.6       & 73.8$\pm$0.9           & 93.4$\pm$0.3          \\
                          & 10                     & 69.8$\pm$0.8       & 72.8$\pm$0.4           & 90.8$\pm$0.2          \\
                          & 15                     & 67.5$\pm$0.5       & 70.1$\pm$1.2           & 88.7$\pm$0.3          \\
                          & 20                     & 64.1$\pm$0.9       & 68.5$\pm$1.0           & 88.0$\pm$0.3          \\
                          & 25                     & 57.0$\pm$3.6       & 64.8$\pm$2.1           & 89.9$\pm$0.5         \\ \bottomrule
\end{tabular}
}
\caption{Effectiveness comparison between \gasoline~and \gasoline-LR}
\label{tab:performance comparison between original and low-rank gasoline}
\end{table}

\noindent\textbf{B - Bilevel Optimization}
Bilevel optimization problem is a powerful mathematical tool with broad applications. For instance, Finn et al.~\cite{finn2017model} formulate the learning to initialization problem in the bilevel optimization context; Li et al.~\cite{li2016data} propose a bilevel optimization-based poisoning attack method for factorization-based systems. There are effective solutions such as the forward and reverse gradients~\cite{franceschi2017forward}, truncated back-propagation~\cite{shaban2019truncated} and so on. In addition, Colson et al.~\cite{colson2007overview} provide a detailed review about this topic. The most related arts to our work are~\cite{chen2020trading} and~\cite{franceschi2019learning}. Both of them aim to modify (or generate from scratch) the given data in a bilevel optimization context. The former studies a data debugging problem under the collaborative filtering scenario whose lower-level problem has a closed-form solution. The latter models every edge with a Bernoulli random variable. As a comparison, the lower-level problem of graph sanitation may not necessarily have a closed-form solution and we modify the initially provided graphs deterministically with versatile variants and broader applications. 


\section{Conclusion}
\label{sec:conclusion}
In this paper, we introduce the {graph sanitation} problem, which aims to improve an initially-provided graph for a given graph mining model. We formulate the graph sanitation problem as a bilevel optimization problem and show that it can be instantiated by a variety of graph mining models such as supervised PageRank, supervised clustering and node classification.  
We further propose an effective solver named \gasoline\ for the graph sanitation problem with semi-supervised node classification. \gasoline\ adopts an efficient approximation of hyper-gradient to guide the modification over the initially-provided graph. \gasoline\ is versatile, and equipped with multiple variants. 
The extensive experimental evaluations demonstrate the broad applicability and effectiveness of the proposed \gasoline. 


\clearpage

\bibliographystyle{ACM-Reference-Format}
\bibliography{www22}


\begin{thebibliography}{47}


\ifx \showCODEN    \undefined \def \showCODEN     #1{\unskip}     \fi
\ifx \showDOI      \undefined \def \showDOI       #1{#1}\fi
\ifx \showISBNx    \undefined \def \showISBNx     #1{\unskip}     \fi
\ifx \showISBNxiii \undefined \def \showISBNxiii  #1{\unskip}     \fi
\ifx \showISSN     \undefined \def \showISSN      #1{\unskip}     \fi
\ifx \showLCCN     \undefined \def \showLCCN      #1{\unskip}     \fi
\ifx \shownote     \undefined \def \shownote      #1{#1}          \fi
\ifx \showarticletitle \undefined \def \showarticletitle #1{#1}   \fi
\ifx \showURL      \undefined \def \showURL       {\relax}        \fi
\providecommand\bibfield[2]{#2}
\providecommand\bibinfo[2]{#2}
\providecommand\natexlab[1]{#1}
\providecommand\showeprint[2][]{arXiv:#2}

\bibitem[\protect\citeauthoryear{Alper, Bach, Henry~Riche, Isenberg, and
  Fekete}{Alper et~al\mbox{.}}{2013}]%
        {alper2013weighted}
\bibfield{author}{\bibinfo{person}{Basak Alper}, \bibinfo{person}{Benjamin
  Bach}, \bibinfo{person}{Nathalie Henry~Riche}, \bibinfo{person}{Tobias
  Isenberg}, {and} \bibinfo{person}{Jean-Daniel Fekete}.}
  \bibinfo{year}{2013}\natexlab{}.
\newblock \showarticletitle{Weighted graph comparison techniques for brain
  connectivity analysis}. In \bibinfo{booktitle}{\emph{SIGCHI}}.
  \bibinfo{pages}{483--492}.
\newblock


\bibitem[\protect\citeauthoryear{Backstrom and Leskovec}{Backstrom and
  Leskovec}{2011}]%
        {backstrom2011supervised}
\bibfield{author}{\bibinfo{person}{Lars Backstrom} {and} \bibinfo{person}{Jure
  Leskovec}.} \bibinfo{year}{2011}\natexlab{}.
\newblock \showarticletitle{Supervised random walks: predicting and
  recommending links in social networks}. In \bibinfo{booktitle}{\emph{WSDM}}.
  \bibinfo{pages}{635--644}.
\newblock


\bibitem[\protect\citeauthoryear{Baydin, Pearlmutter, Radul, and
  Siskind}{Baydin et~al\mbox{.}}{2017}]%
        {baydin2017automatic}
\bibfield{author}{\bibinfo{person}{At{\i}l{\i}m~G{\"u}nes Baydin},
  \bibinfo{person}{Barak~A Pearlmutter}, \bibinfo{person}{Alexey~Andreyevich
  Radul}, {and} \bibinfo{person}{Jeffrey~Mark Siskind}.}
  \bibinfo{year}{2017}\natexlab{}.
\newblock \showarticletitle{Automatic differentiation in machine learning: a
  survey}.
\newblock \bibinfo{journal}{\emph{The Journal of Machine Learning Research}}
  \bibinfo{volume}{18}, \bibinfo{number}{1} (\bibinfo{year}{2017}),
  \bibinfo{pages}{5595--5637}.
\newblock


\bibitem[\protect\citeauthoryear{Bordes, Usunier, Garcia-Duran, Weston, and
  Yakhnenko}{Bordes et~al\mbox{.}}{2013}]%
        {bordes2013translating}
\bibfield{author}{\bibinfo{person}{Antoine Bordes}, \bibinfo{person}{Nicolas
  Usunier}, \bibinfo{person}{Alberto Garcia-Duran}, \bibinfo{person}{Jason
  Weston}, {and} \bibinfo{person}{Oksana Yakhnenko}.}
  \bibinfo{year}{2013}\natexlab{}.
\newblock \showarticletitle{Translating embeddings for modeling
  multi-relational data}. In \bibinfo{booktitle}{\emph{NIPS}}.
  \bibinfo{pages}{1--9}.
\newblock


\bibitem[\protect\citeauthoryear{Chen, Peng, Ying, and Tong}{Chen
  et~al\mbox{.}}{2018}]%
        {chen2018network}
\bibfield{author}{\bibinfo{person}{Chen Chen}, \bibinfo{person}{Ruiyue Peng},
  \bibinfo{person}{Lei Ying}, {and} \bibinfo{person}{Hanghang Tong}.}
  \bibinfo{year}{2018}\natexlab{}.
\newblock \showarticletitle{Network connectivity optimization: Fundamental
  limits and effective algorithms}. In \bibinfo{booktitle}{\emph{SIGKDD}}.
  \bibinfo{pages}{1167--1176}.
\newblock


\bibitem[\protect\citeauthoryear{Chen, Tong, Prakash, Tsourakakis, Eliassi-Rad,
  Faloutsos, and Chau}{Chen et~al\mbox{.}}{2015}]%
        {chen2015node}
\bibfield{author}{\bibinfo{person}{Chen Chen}, \bibinfo{person}{Hanghang Tong},
  \bibinfo{person}{B~Aditya Prakash}, \bibinfo{person}{Charalampos~E
  Tsourakakis}, \bibinfo{person}{Tina Eliassi-Rad}, \bibinfo{person}{Christos
  Faloutsos}, {and} \bibinfo{person}{Duen~Horng Chau}.}
  \bibinfo{year}{2015}\natexlab{}.
\newblock \showarticletitle{Node immunization on large graphs: Theory and
  algorithms}.
\newblock \bibinfo{journal}{\emph{IEEE Transactions on Knowledge and Data
  Engineering}} \bibinfo{volume}{28}, \bibinfo{number}{1}
  (\bibinfo{year}{2015}), \bibinfo{pages}{113--126}.
\newblock


\bibitem[\protect\citeauthoryear{Chen, Yao, Xu, Xu, and Tong}{Chen
  et~al\mbox{.}}{2020}]%
        {chen2020trading}
\bibfield{author}{\bibinfo{person}{Long Chen}, \bibinfo{person}{Yuan Yao},
  \bibinfo{person}{Feng Xu}, \bibinfo{person}{Miao Xu}, {and}
  \bibinfo{person}{Hanghang Tong}.} \bibinfo{year}{2020}\natexlab{}.
\newblock \showarticletitle{Trading Personalization for Accuracy: Data
  Debugging in Collaborative Filtering}.
\newblock \bibinfo{journal}{\emph{NeurIPS}}  \bibinfo{volume}{33}
  (\bibinfo{year}{2020}).
\newblock


\bibitem[\protect\citeauthoryear{Colson, Marcotte, and Savard}{Colson
  et~al\mbox{.}}{2007}]%
        {colson2007overview}
\bibfield{author}{\bibinfo{person}{Beno{\^\i}t Colson},
  \bibinfo{person}{Patrice Marcotte}, {and} \bibinfo{person}{Gilles Savard}.}
  \bibinfo{year}{2007}\natexlab{}.
\newblock \showarticletitle{An overview of bilevel optimization}.
\newblock \bibinfo{journal}{\emph{Annals of operations research}}
  \bibinfo{volume}{153}, \bibinfo{number}{1} (\bibinfo{year}{2007}),
  \bibinfo{pages}{235--256}.
\newblock


\bibitem[\protect\citeauthoryear{Entezari, Al-Sayouri, Darvishzadeh, and
  Papalexakis}{Entezari et~al\mbox{.}}{2020}]%
        {entezari2020all}
\bibfield{author}{\bibinfo{person}{Negin Entezari}, \bibinfo{person}{Saba~A
  Al-Sayouri}, \bibinfo{person}{Amirali Darvishzadeh}, {and}
  \bibinfo{person}{Evangelos~E Papalexakis}.} \bibinfo{year}{2020}\natexlab{}.
\newblock \showarticletitle{All You Need Is Low (Rank) Defending Against
  Adversarial Attacks on Graphs}. In \bibinfo{booktitle}{\emph{WSDM}}.
  \bibinfo{pages}{169--177}.
\newblock


\bibitem[\protect\citeauthoryear{Finn, Abbeel, and Levine}{Finn
  et~al\mbox{.}}{2017}]%
        {finn2017model}
\bibfield{author}{\bibinfo{person}{Chelsea Finn}, \bibinfo{person}{Pieter
  Abbeel}, {and} \bibinfo{person}{Sergey Levine}.}
  \bibinfo{year}{2017}\natexlab{}.
\newblock \showarticletitle{Model-agnostic meta-learning for fast adaptation of
  deep networks}. In \bibinfo{booktitle}{\emph{ICML}}. PMLR,
  \bibinfo{pages}{1126--1135}.
\newblock


\bibitem[\protect\citeauthoryear{Franceschi, Donini, Frasconi, and
  Pontil}{Franceschi et~al\mbox{.}}{2017}]%
        {franceschi2017forward}
\bibfield{author}{\bibinfo{person}{Luca Franceschi}, \bibinfo{person}{Michele
  Donini}, \bibinfo{person}{Paolo Frasconi}, {and}
  \bibinfo{person}{Massimiliano Pontil}.} \bibinfo{year}{2017}\natexlab{}.
\newblock \showarticletitle{Forward and reverse gradient-based hyperparameter
  optimization}. In \bibinfo{booktitle}{\emph{ICML}}. PMLR,
  \bibinfo{pages}{1165--1173}.
\newblock


\bibitem[\protect\citeauthoryear{Franceschi, Niepert, Pontil, and
  He}{Franceschi et~al\mbox{.}}{2019}]%
        {franceschi2019learning}
\bibfield{author}{\bibinfo{person}{Luca Franceschi}, \bibinfo{person}{Mathias
  Niepert}, \bibinfo{person}{Massimiliano Pontil}, {and} \bibinfo{person}{Xiao
  He}.} \bibinfo{year}{2019}\natexlab{}.
\newblock \showarticletitle{Learning discrete structures for graph neural
  networks}. In \bibinfo{booktitle}{\emph{ICML}}. PMLR,
  \bibinfo{pages}{1972--1982}.
\newblock


\bibitem[\protect\citeauthoryear{Gyongyi, Garcia-Molina, and Pedersen}{Gyongyi
  et~al\mbox{.}}{2004}]%
        {gyongyi2004combating}
\bibfield{author}{\bibinfo{person}{Zoltan Gyongyi}, \bibinfo{person}{Hector
  Garcia-Molina}, {and} \bibinfo{person}{Jan Pedersen}.}
  \bibinfo{year}{2004}\natexlab{}.
\newblock \showarticletitle{Combating web spam with trustrank}. In
  \bibinfo{booktitle}{\emph{VLDB}}.
\newblock


\bibitem[\protect\citeauthoryear{Haveliwala}{Haveliwala}{2003}]%
        {haveliwala2003topic}
\bibfield{author}{\bibinfo{person}{Taher~H Haveliwala}.}
  \bibinfo{year}{2003}\natexlab{}.
\newblock \showarticletitle{Topic-sensitive pagerank: A context-sensitive
  ranking algorithm for web search}.
\newblock \bibinfo{journal}{\emph{IEEE transactions on knowledge and data
  engineering}} \bibinfo{volume}{15}, \bibinfo{number}{4}
  (\bibinfo{year}{2003}), \bibinfo{pages}{784--796}.
\newblock


\bibitem[\protect\citeauthoryear{Huisman}{Huisman}{2009}]%
        {huisman2009imputation}
\bibfield{author}{\bibinfo{person}{Mark Huisman}.}
  \bibinfo{year}{2009}\natexlab{}.
\newblock \showarticletitle{Imputation of missing network data: Some simple
  procedures}.
\newblock \bibinfo{journal}{\emph{Journal of Social Structure}}
  \bibinfo{volume}{10}, \bibinfo{number}{1} (\bibinfo{year}{2009}),
  \bibinfo{pages}{1--29}.
\newblock


\bibitem[\protect\citeauthoryear{Jeh and Widom}{Jeh and Widom}{2003}]%
        {jeh2003scaling}
\bibfield{author}{\bibinfo{person}{Glen Jeh} {and} \bibinfo{person}{Jennifer
  Widom}.} \bibinfo{year}{2003}\natexlab{}.
\newblock \showarticletitle{Scaling personalized web search}. In
  \bibinfo{booktitle}{\emph{WWW}}. \bibinfo{pages}{271--279}.
\newblock


\bibitem[\protect\citeauthoryear{Jin, Ma, Liu, Tang, Wang, and Tang}{Jin
  et~al\mbox{.}}{2020}]%
        {DBLP:conf/kdd/Jin0LTWT20}
\bibfield{author}{\bibinfo{person}{Wei Jin}, \bibinfo{person}{Yao Ma},
  \bibinfo{person}{Xiaorui Liu}, \bibinfo{person}{Xianfeng Tang},
  \bibinfo{person}{Suhang Wang}, {and} \bibinfo{person}{Jiliang Tang}.}
  \bibinfo{year}{2020}\natexlab{}.
\newblock \showarticletitle{Graph Structure Learning for Robust Graph Neural
  Networks}. In \bibinfo{booktitle}{\emph{SIGKDD}}. \bibinfo{publisher}{{ACM}},
  \bibinfo{pages}{66--74}.
\newblock


\bibitem[\protect\citeauthoryear{Kang and Tong}{Kang and Tong}{2019}]%
        {kang2019n2n}
\bibfield{author}{\bibinfo{person}{Jian Kang} {and} \bibinfo{person}{Hanghang
  Tong}.} \bibinfo{year}{2019}\natexlab{}.
\newblock \showarticletitle{N2n: Network derivative mining}. In
  \bibinfo{booktitle}{\emph{CIKM}}. \bibinfo{pages}{861--870}.
\newblock


\bibitem[\protect\citeauthoryear{Keeling and Eames}{Keeling and Eames}{2005}]%
        {keeling2005networks}
\bibfield{author}{\bibinfo{person}{Matt~J Keeling} {and}
  \bibinfo{person}{Ken~TD Eames}.} \bibinfo{year}{2005}\natexlab{}.
\newblock \showarticletitle{Networks and epidemic models}.
\newblock \bibinfo{journal}{\emph{Journal of the Royal Society Interface}}
  \bibinfo{volume}{2}, \bibinfo{number}{4} (\bibinfo{year}{2005}),
  \bibinfo{pages}{295--307}.
\newblock


\bibitem[\protect\citeauthoryear{Kipf and Welling}{Kipf and Welling}{2017}]%
        {kipf2016semi}
\bibfield{author}{\bibinfo{person}{Thomas~N. Kipf} {and} \bibinfo{person}{Max
  Welling}.} \bibinfo{year}{2017}\natexlab{}.
\newblock \showarticletitle{Semi-Supervised Classification with Graph
  Convolutional Networks}. In \bibinfo{booktitle}{\emph{ICLR}}.
\newblock


\bibitem[\protect\citeauthoryear{Klicpera, Bojchevski, and
  G{\"u}nnemann}{Klicpera et~al\mbox{.}}{2018}]%
        {klicpera2018predict}
\bibfield{author}{\bibinfo{person}{Johannes Klicpera},
  \bibinfo{person}{Aleksandar Bojchevski}, {and} \bibinfo{person}{Stephan
  G{\"u}nnemann}.} \bibinfo{year}{2018}\natexlab{}.
\newblock \showarticletitle{Predict then Propagate: Graph Neural Networks meet
  Personalized PageRank}. In \bibinfo{booktitle}{\emph{ICLR}}.
\newblock


\bibitem[\protect\citeauthoryear{Li, Wang, Singh, and Vorobeychik}{Li
  et~al\mbox{.}}{2016a}]%
        {li2016data}
\bibfield{author}{\bibinfo{person}{Bo Li}, \bibinfo{person}{Yining Wang},
  \bibinfo{person}{Aarti Singh}, {and} \bibinfo{person}{Yevgeniy Vorobeychik}.}
  \bibinfo{year}{2016}\natexlab{a}.
\newblock \showarticletitle{Data poisoning attacks on factorization-based
  collaborative filtering}. In \bibinfo{booktitle}{\emph{NIPS}}.
  \bibinfo{pages}{1893--1901}.
\newblock


\bibitem[\protect\citeauthoryear{Li, Liu, Zhang, Wang, Wen, Pan, and Cheng}{Li
  et~al\mbox{.}}{2021}]%
        {li2021mask}
\bibfield{author}{\bibinfo{person}{Jia Li}, \bibinfo{person}{Mengzhou Liu},
  \bibinfo{person}{Honglei Zhang}, \bibinfo{person}{Pengyun Wang},
  \bibinfo{person}{Yong Wen}, \bibinfo{person}{Lujia Pan}, {and}
  \bibinfo{person}{Hong Cheng}.} \bibinfo{year}{2021}\natexlab{}.
\newblock \showarticletitle{Mask-GVAE: Blind Denoising Graphs via Partition}.
  In \bibinfo{booktitle}{\emph{TheWebConf}}. \bibinfo{pages}{3688--3698}.
\newblock


\bibitem[\protect\citeauthoryear{Li, Yao, Tang, Fan, and Tong}{Li
  et~al\mbox{.}}{2016b}]%
        {li2016quint}
\bibfield{author}{\bibinfo{person}{Liangyue Li}, \bibinfo{person}{Yuan Yao},
  \bibinfo{person}{Jie Tang}, \bibinfo{person}{Wei Fan}, {and}
  \bibinfo{person}{Hanghang Tong}.} \bibinfo{year}{2016}\natexlab{b}.
\newblock \showarticletitle{QUINT: on query-specific optimal networks}. In
  \bibinfo{booktitle}{\emph{SIGKDD}}. \bibinfo{pages}{985--994}.
\newblock


\bibitem[\protect\citeauthoryear{Liben-Nowell and Kleinberg}{Liben-Nowell and
  Kleinberg}{2007}]%
        {liben2007link}
\bibfield{author}{\bibinfo{person}{David Liben-Nowell} {and}
  \bibinfo{person}{Jon Kleinberg}.} \bibinfo{year}{2007}\natexlab{}.
\newblock \showarticletitle{The link-prediction problem for social networks}.
\newblock \bibinfo{journal}{\emph{Journal of the American society for
  information science and technology}} \bibinfo{volume}{58},
  \bibinfo{number}{7} (\bibinfo{year}{2007}), \bibinfo{pages}{1019--1031}.
\newblock


\bibitem[\protect\citeauthoryear{Nichol, Achiam, and Schulman}{Nichol
  et~al\mbox{.}}{2018}]%
        {nichol2018first}
\bibfield{author}{\bibinfo{person}{Alex Nichol}, \bibinfo{person}{Joshua
  Achiam}, {and} \bibinfo{person}{John Schulman}.}
  \bibinfo{year}{2018}\natexlab{}.
\newblock \showarticletitle{On first-order meta-learning algorithms}.
\newblock \bibinfo{journal}{\emph{arXiv preprint arXiv:1803.02999}}
  (\bibinfo{year}{2018}).
\newblock


\bibitem[\protect\citeauthoryear{Page, Brin, Motwani, and Winograd}{Page
  et~al\mbox{.}}{1999}]%
        {page1999pagerank}
\bibfield{author}{\bibinfo{person}{Lawrence Page}, \bibinfo{person}{Sergey
  Brin}, \bibinfo{person}{Rajeev Motwani}, {and} \bibinfo{person}{Terry
  Winograd}.} \bibinfo{year}{1999}\natexlab{}.
\newblock \bibinfo{booktitle}{\emph{The PageRank citation ranking: Bringing
  order to the web.}}
\newblock \bibinfo{type}{{T}echnical {R}eport}. \bibinfo{institution}{Stanford
  InfoLab}.
\newblock


\bibitem[\protect\citeauthoryear{Rong, Huang, Xu, and Huang}{Rong
  et~al\mbox{.}}{2019}]%
        {rong2019dropedge}
\bibfield{author}{\bibinfo{person}{Yu Rong}, \bibinfo{person}{Wenbing Huang},
  \bibinfo{person}{Tingyang Xu}, {and} \bibinfo{person}{Junzhou Huang}.}
  \bibinfo{year}{2019}\natexlab{}.
\newblock \showarticletitle{DropEdge: Towards Deep Graph Convolutional Networks
  on Node Classification}. In \bibinfo{booktitle}{\emph{ICLR}}.
\newblock


\bibitem[\protect\citeauthoryear{Shaban, Cheng, Hatch, and Boots}{Shaban
  et~al\mbox{.}}{2019}]%
        {shaban2019truncated}
\bibfield{author}{\bibinfo{person}{Amirreza Shaban}, \bibinfo{person}{Ching-An
  Cheng}, \bibinfo{person}{Nathan Hatch}, {and} \bibinfo{person}{Byron Boots}.}
  \bibinfo{year}{2019}\natexlab{}.
\newblock \showarticletitle{Truncated back-propagation for bilevel
  optimization}. In \bibinfo{booktitle}{\emph{AISTATS}}. PMLR,
  \bibinfo{pages}{1723--1732}.
\newblock


\bibitem[\protect\citeauthoryear{Shi and Malik}{Shi and Malik}{2000}]%
        {shi2000normalized}
\bibfield{author}{\bibinfo{person}{Jianbo Shi} {and} \bibinfo{person}{Jitendra
  Malik}.} \bibinfo{year}{2000}\natexlab{}.
\newblock \showarticletitle{Normalized cuts and image segmentation}.
\newblock \bibinfo{journal}{\emph{IEEE Transactions on pattern analysis and
  machine intelligence}} \bibinfo{volume}{22}, \bibinfo{number}{8}
  (\bibinfo{year}{2000}), \bibinfo{pages}{888--905}.
\newblock


\bibitem[\protect\citeauthoryear{Tong, Faloutsos, and Pan}{Tong
  et~al\mbox{.}}{2006}]%
        {tong2006fast}
\bibfield{author}{\bibinfo{person}{Hanghang Tong}, \bibinfo{person}{Christos
  Faloutsos}, {and} \bibinfo{person}{Jia-Yu Pan}.}
  \bibinfo{year}{2006}\natexlab{}.
\newblock \showarticletitle{Fast random walk with restart and its
  applications}. In \bibinfo{booktitle}{\emph{ICDM}}. IEEE,
  \bibinfo{pages}{613--622}.
\newblock


\bibitem[\protect\citeauthoryear{Trouillon, Welbl, Riedel, Gaussier, and
  Bouchard}{Trouillon et~al\mbox{.}}{2016}]%
        {trouillon2016complex}
\bibfield{author}{\bibinfo{person}{Th{\'e}o Trouillon},
  \bibinfo{person}{Johannes Welbl}, \bibinfo{person}{Sebastian Riedel},
  \bibinfo{person}{{\'E}ric Gaussier}, {and} \bibinfo{person}{Guillaume
  Bouchard}.} \bibinfo{year}{2016}\natexlab{}.
\newblock \showarticletitle{Complex embeddings for simple link prediction}. In
  \bibinfo{booktitle}{\emph{ICML}}. PMLR, \bibinfo{pages}{2071--2080}.
\newblock


\bibitem[\protect\citeauthoryear{Van Der~Maaten}{Van Der~Maaten}{2014}]%
        {van2014accelerating}
\bibfield{author}{\bibinfo{person}{Laurens Van Der~Maaten}.}
  \bibinfo{year}{2014}\natexlab{}.
\newblock \showarticletitle{Accelerating t-SNE using tree-based algorithms}.
\newblock \bibinfo{journal}{\emph{The Journal of Machine Learning Research}}
  \bibinfo{volume}{15}, \bibinfo{number}{1} (\bibinfo{year}{2014}),
  \bibinfo{pages}{3221--3245}.
\newblock


\bibitem[\protect\citeauthoryear{Veli{\v{c}}kovi{\'c}, Cucurull, Casanova,
  Romero, Li{\`o}, and Bengio}{Veli{\v{c}}kovi{\'c} et~al\mbox{.}}{2018}]%
        {velivckovic2018graph}
\bibfield{author}{\bibinfo{person}{Petar Veli{\v{c}}kovi{\'c}},
  \bibinfo{person}{Guillem Cucurull}, \bibinfo{person}{Arantxa Casanova},
  \bibinfo{person}{Adriana Romero}, \bibinfo{person}{Pietro Li{\`o}}, {and}
  \bibinfo{person}{Yoshua Bengio}.} \bibinfo{year}{2018}\natexlab{}.
\newblock \showarticletitle{Graph Attention Networks}. In
  \bibinfo{booktitle}{\emph{ICLR}}.
\newblock


\bibitem[\protect\citeauthoryear{Wagstaff and Cardie}{Wagstaff and
  Cardie}{2000}]%
        {wagstaff2000clustering}
\bibfield{author}{\bibinfo{person}{Kiri Wagstaff} {and} \bibinfo{person}{Claire
  Cardie}.} \bibinfo{year}{2000}\natexlab{}.
\newblock \showarticletitle{Clustering with instance-level constraints}.
\newblock \bibinfo{journal}{\emph{AAAI/IAAI}}  \bibinfo{volume}{1097}
  (\bibinfo{year}{2000}), \bibinfo{pages}{577--584}.
\newblock


\bibitem[\protect\citeauthoryear{Wang, Pourshafeie, Zitnik, Zhu, Bustamante,
  Batzoglou, and Leskovec}{Wang et~al\mbox{.}}{2018}]%
        {wang2018network}
\bibfield{author}{\bibinfo{person}{Bo Wang}, \bibinfo{person}{Armin
  Pourshafeie}, \bibinfo{person}{Marinka Zitnik}, \bibinfo{person}{Junjie Zhu},
  \bibinfo{person}{Carlos~D Bustamante}, \bibinfo{person}{Serafim Batzoglou},
  {and} \bibinfo{person}{Jure Leskovec}.} \bibinfo{year}{2018}\natexlab{}.
\newblock \showarticletitle{Network enhancement as a general method to denoise
  weighted biological networks}.
\newblock \bibinfo{journal}{\emph{Nature communications}} \bibinfo{volume}{9},
  \bibinfo{number}{1} (\bibinfo{year}{2018}), \bibinfo{pages}{1--8}.
\newblock


\bibitem[\protect\citeauthoryear{Wang, Lin, Cui, Jia, Wang, Fang, Yu, Zhou,
  Yang, and Qi}{Wang et~al\mbox{.}}{2019}]%
        {wang2019semi}
\bibfield{author}{\bibinfo{person}{Daixin Wang}, \bibinfo{person}{Jianbin Lin},
  \bibinfo{person}{Peng Cui}, \bibinfo{person}{Quanhui Jia},
  \bibinfo{person}{Zhen Wang}, \bibinfo{person}{Yanming Fang},
  \bibinfo{person}{Quan Yu}, \bibinfo{person}{Jun Zhou},
  \bibinfo{person}{Shuang Yang}, {and} \bibinfo{person}{Yuan Qi}.}
  \bibinfo{year}{2019}\natexlab{}.
\newblock \showarticletitle{A semi-supervised graph attentive network for
  financial fraud detection}. In \bibinfo{booktitle}{\emph{ICDM}}. IEEE,
  \bibinfo{pages}{598--607}.
\newblock


\bibitem[\protect\citeauthoryear{Wang and Davidson}{Wang and Davidson}{2010}]%
        {wang2010flexible}
\bibfield{author}{\bibinfo{person}{Xiang Wang} {and} \bibinfo{person}{Ian
  Davidson}.} \bibinfo{year}{2010}\natexlab{}.
\newblock \showarticletitle{Flexible constrained spectral clustering}. In
  \bibinfo{booktitle}{\emph{SIGKDD}}. \bibinfo{pages}{563--572}.
\newblock


\bibitem[\protect\citeauthoryear{Wang, Zhang, Feng, and Chen}{Wang
  et~al\mbox{.}}{2014}]%
        {wang2014knowledge}
\bibfield{author}{\bibinfo{person}{Zhen Wang}, \bibinfo{person}{Jianwen Zhang},
  \bibinfo{person}{Jianlin Feng}, {and} \bibinfo{person}{Zheng Chen}.}
  \bibinfo{year}{2014}\natexlab{}.
\newblock \showarticletitle{Knowledge graph embedding by translating on
  hyperplanes}. In \bibinfo{booktitle}{\emph{AAAI}}, Vol.~\bibinfo{volume}{28}.
\newblock


\bibitem[\protect\citeauthoryear{Wu, Souza, Zhang, Fifty, Yu, and
  Weinberger}{Wu et~al\mbox{.}}{2019a}]%
        {wu2019simplifying}
\bibfield{author}{\bibinfo{person}{Felix Wu}, \bibinfo{person}{Amauri Souza},
  \bibinfo{person}{Tianyi Zhang}, \bibinfo{person}{Christopher Fifty},
  \bibinfo{person}{Tao Yu}, {and} \bibinfo{person}{Kilian Weinberger}.}
  \bibinfo{year}{2019}\natexlab{a}.
\newblock \showarticletitle{Simplifying Graph Convolutional Networks}. In
  \bibinfo{booktitle}{\emph{ICML}}. \bibinfo{pages}{6861--6871}.
\newblock


\bibitem[\protect\citeauthoryear{Wu, Wang, Tyshetskiy, Docherty, Lu, and
  Zhu}{Wu et~al\mbox{.}}{2019b}]%
        {wu2019adversarial}
\bibfield{author}{\bibinfo{person}{Huijun Wu}, \bibinfo{person}{Chen Wang},
  \bibinfo{person}{Yuriy Tyshetskiy}, \bibinfo{person}{Andrew Docherty},
  \bibinfo{person}{Kai Lu}, {and} \bibinfo{person}{Liming Zhu}.}
  \bibinfo{year}{2019}\natexlab{b}.
\newblock \showarticletitle{Adversarial examples for graph data: deep insights
  into attack and defense}. In \bibinfo{booktitle}{\emph{IJCAI}}. AAAI Press,
  \bibinfo{pages}{4816--4823}.
\newblock


\bibitem[\protect\citeauthoryear{Xu, Yang, Wang, Liu, Zhang, Chen, and Lu}{Xu
  et~al\mbox{.}}{2020}]%
        {xu2020robust}
\bibfield{author}{\bibinfo{person}{Jiarong Xu}, \bibinfo{person}{Yang Yang},
  \bibinfo{person}{Chunping Wang}, \bibinfo{person}{Zongtao Liu},
  \bibinfo{person}{Jing Zhang}, \bibinfo{person}{Lei Chen}, {and}
  \bibinfo{person}{Jiangang Lu}.} \bibinfo{year}{2020}\natexlab{}.
\newblock \showarticletitle{Robust Network Enhancement from Flawed Networks}.
\newblock \bibinfo{journal}{\emph{IEEE Transactions on Knowledge and Data
  Engineering}} (\bibinfo{year}{2020}).
\newblock


\bibitem[\protect\citeauthoryear{Yan, Dodier, Mozer, and Wolniewicz}{Yan
  et~al\mbox{.}}{2003}]%
        {yan2003optimizing}
\bibfield{author}{\bibinfo{person}{Lian Yan}, \bibinfo{person}{Robert Dodier},
  \bibinfo{person}{Michael~C Mozer}, {and} \bibinfo{person}{Richard
  Wolniewicz}.} \bibinfo{year}{2003}\natexlab{}.
\newblock \showarticletitle{Optimizing classifier performance via an
  approximation to the Wilcoxon-Mann-Whitney statistic}. In
  \bibinfo{booktitle}{\emph{ICML}}. \bibinfo{pages}{848--855}.
\newblock


\bibitem[\protect\citeauthoryear{Zafarani, Abbasi, and Liu}{Zafarani
  et~al\mbox{.}}{2014}]%
        {zafarani2014social}
\bibfield{author}{\bibinfo{person}{Reza Zafarani},
  \bibinfo{person}{Mohammad~Ali Abbasi}, {and} \bibinfo{person}{Huan Liu}.}
  \bibinfo{year}{2014}\natexlab{}.
\newblock \bibinfo{booktitle}{\emph{Social media mining: an introduction}}.
\newblock \bibinfo{publisher}{Cambridge University Press}.
\newblock


\bibitem[\protect\citeauthoryear{Zhu, Zhang, Cui, and Zhu}{Zhu
  et~al\mbox{.}}{2019}]%
        {zhu2019robust}
\bibfield{author}{\bibinfo{person}{Dingyuan Zhu}, \bibinfo{person}{Ziwei
  Zhang}, \bibinfo{person}{Peng Cui}, {and} \bibinfo{person}{Wenwu Zhu}.}
  \bibinfo{year}{2019}\natexlab{}.
\newblock \showarticletitle{Robust graph convolutional networks against
  adversarial attacks}. In \bibinfo{booktitle}{\emph{SIGKDD}}.
  \bibinfo{pages}{1399--1407}.
\newblock


\bibitem[\protect\citeauthoryear{Z{\"u}gner, Akbarnejad, and
  G{\"u}nnemann}{Z{\"u}gner et~al\mbox{.}}{2018}]%
        {zugner2018adversarial}
\bibfield{author}{\bibinfo{person}{Daniel Z{\"u}gner}, \bibinfo{person}{Amir
  Akbarnejad}, {and} \bibinfo{person}{Stephan G{\"u}nnemann}.}
  \bibinfo{year}{2018}\natexlab{}.
\newblock \showarticletitle{Adversarial attacks on neural networks for graph
  data}. In \bibinfo{booktitle}{\emph{SIGKDD}}. \bibinfo{pages}{2847--2856}.
\newblock


\bibitem[\protect\citeauthoryear{Z{\"{u}}gner and G{\"{u}}nnemann}{Z{\"{u}}gner
  and G{\"{u}}nnemann}{2019}]%
        {DBLP:conf/iclr/ZugnerG19}
\bibfield{author}{\bibinfo{person}{Daniel Z{\"{u}}gner} {and}
  \bibinfo{person}{Stephan G{\"{u}}nnemann}.} \bibinfo{year}{2019}\natexlab{}.
\newblock \showarticletitle{Adversarial Attacks on Graph Neural Networks via
  Meta Learning}. In \bibinfo{booktitle}{\emph{ICLR}}.
\newblock


\end{thebibliography}

\clearpage

\appendix

\section{Reproducibility}
\label{sec:reproducibility}
We will release the source code upon the publication of the paper. Three public graph datasets, Cora\footnote{\label{fn:gcn}https://github.com/tkipf/gcn}, Citeseer\footref{fn:gcn}, and Polbolgs~\footnote{https://github.com/ChandlerBang/Pro-GNN}, are used in our paper with following detailed statistics in Table~\ref{tab:dataset statistics}

\vspace{-3mm}

\begin{table}[ht]
\resizebox{0.8\linewidth}{!}{
\begin{tabular}{c|cccc}
\toprule
\textbf{Data} & \textbf{Nodes} & \textbf{Edges} & \textbf{Classes} & \textbf{Features} \\ \midrule
Cora          & 2,485        & 5,069        & 7                & 1,433             \\
Citeseer      & 2,110        & 3,668        & 6                & 3,703             \\
Polblogs      & 1,222        & 16,714       & 2                & N/A               \\ \bottomrule
\end{tabular}
}
\caption{Statistics of Datasets}
\label{tab:dataset statistics}
\end{table}

\vspace{-12mm}

\subsection{Hyper-Parameter Settings}
We summarize the hyper-parameter settings of models implemented in our experiments including \emph{baseline methods}, \emph{backbone} and \emph{downstream classifiers} of \gasoline:
\begin{itemize}[noitemsep,leftmargin=9pt]
    \item \textbf{GCN~\cite{kipf2016semi}/GAT~\cite{velivckovic2018graph}/RGCN~\cite{zhu2019robust}:} We follow the default settings of publicly available implementation of GCN\footref{fn:gcn}, GAT\footnote{https://github.com/PetarV-/GAT}, RGCN\footnote{https://github.com/ZW-ZHANG/RobustGCN}.
    
    \item \textbf{SGC~\cite{wu2019simplifying}:} The implementation of SGC is the same as GCN~\cite{kipf2016semi} but we remove the activation function of the hidden layers.
    
    \item \textbf{APPNP~\cite{klicpera2018predict}:} We follow the recommended hyper-parameter settings of APPNP~\cite{klicpera2018predict} in the original paper.
    
    
    \item \textbf{Jaccard~\cite{wu2019adversarial}: } We search the edge removing threshold of Jaccard similarity from $\{0.01, 0.03, 0.05, 0.1, 0.3, 0.5\}$ and report the best.
    
    \item \textbf{SVD~\cite{entezari2020all}: } We search the rank of SVD from $\{10,30,50,100\\,200,300\}$ and report the best results from the above settings.
    
    
    \item \textbf{DE~\cite{rong2019dropedge}: } We search the dropout rate of DE over existing edges from $\{0.05, 0.1, 0.15, 0.2, 0.25\}$ and report the best results from the above settings.
    
    \item \textbf{LDS~\cite{franceschi2019learning}: } We implement LDS with the same modification budget as \gasoline-CT.

\end{itemize}

The detailed settings of \gasoline~ are as follows: (1) for all the modification strategies (discretized vs. continuous and topology vs. feature), the modification budget towards topology $B_{\texttt{topo}}$ and the modification budget towards feature $B_{\texttt{fea}}$ are introduced in the Section~\ref{sec:experiment setups}. Specifically, the $\texttt{modification rate}_{\texttt{topo}}=0.1$ and $\texttt{modification rate}_{\texttt{fea}}=0.001$. We modify the graph in $10$ steps so the budget $b$ in every modification step is $\lfloor\frac{B}{10}\rfloor$ (i.e., $b_{\texttt{topo}}=\lfloor\frac{B_{\texttt{topo}}}{10}\rfloor$ and $b_{\texttt{fea}}=\lfloor\frac{B_{\texttt{fea}}}{10}\rfloor$). (2) the settings of backbone classifiers and downstream classifiers (GCN~\cite{kipf2016semi}, SGC~\cite{wu2019simplifying}, APPNP~\cite{klicpera2018predict}) used in our experiments follow the aforementioned settings. (3) the number of iterations for the optimization of lower-level problem $T$ is set as $200$ and the truncating iteration $P$ is set as $196$. The number of folds $K$ is set as $8$.

For the attacking methods, their attacking perturbation rates are introduced in the Section~\ref{sec:effectivenss of gasoline}, and here we present the detailed implementation of them. (1) We follow the publicly-available implementation\footnote{https://github.com/danielzuegner/gnn-meta-attack} of metattack~\cite{DBLP:conf/iclr/ZugnerG19} and adopt the `Meta-Self' variant to attack the provided graphs; (2) we follow \cite{DBLP:conf/kdd/Jin0LTWT20} to select nodes with degree larger than $10$ as the target nodes and implement \nettack~ with the publicly-available implementation\footnote{https://github.com/danielzuegner/nettack}; (3) we implement random attack by symmetrically flipping $B_{\texttt{attack}}$ entries of the adjacency matrix $\mathbf{A}$ of provided graphs.

\section{Proof of Lemma~\ref{lm:time and space complexity}}




Our complexity analysis is mainly based on the propagation formula of matrix multiplication-based GNNs and only focuses on a single-layered GNN with first order approximation of the hyper-gradient. However, it can be easily generalized into a wide range of scenarios (e.g., multi-layered GNNs) with similar analysis and conclusions. For brevity, we only analyze the complexity of computing $\nabla_{\mathbf{U}}\mathcal{L}$. The analysis w.r.t. computing $\nabla_{\mathbf{V}}\mathcal{L}$ is similar and we omit here.



\vspace{-2mm}
\begin{proof}
For typical matrix multiplication-based GNNs (e.g., \cite{kipf2016semi}), their propagation formula can be represented as $\mathbf{X}\leftarrow\sigma(\mathbf{A}\mathbf{X}\mathbf{W})$ (or even simplified by removing the nonlinear activation function $\sigma$ and feature transformation matrix $\mathbf{W}$ between several layers~\cite{wu2019simplifying,klicpera2018predict}). If we do not consider the gradient across the model parameter (i.e., $\mathbf{W}$) updating trajectory (i.e., first order approximation), and assume that our GNN contains only one layer, the hyper-gradient with respect to the vector $\mathbf{U}$ can be computed as follows,

\vspace{-4mm}
\begin{equation}
\label{eq:explicit approximated hyper-gradient}
    \nabla_{\mathbf{U}}\mathcal{L} = [\frac{\partial \mathcal{L}}{\partial \sigma((\mathbf{A}+\mathbf{U}\mathbf{V}^{\prime})\mathbf{X}\mathbf{W})}\circ \sigma^{\prime}((\mathbf{A}+\mathbf{U}\mathbf{V}^{\prime})\mathbf{X}\mathbf{W})]\mathbf{W}^{\prime}\mathbf{X}^{\prime}\mathbf{V}.
\end{equation}

\noindent The computation of $(\mathbf{A}+\mathbf{U}\mathbf{V}^{\prime})\mathbf{X}\mathbf{W}$ can be rewritten as $\mathbf{A}\mathbf{X}\mathbf{W}+\mathbf{U}\mathbf{V}^{\prime}\mathbf{X}\mathbf{W}$. Note that $\mathbf{A}$ is a sparse matrix and the space cost is $O(m)$ for computing $\mathbf{A}\mathbf{X}\mathbf{W}$. The space cost is $O(nd)$ for $\mathbf{U}\mathbf{V}^{\prime}\mathbf{X}\mathbf{W}$. For $\mathbf{W}^{\prime}\mathbf{X}^{\prime}\mathbf{V}$ the space cost is $O(nd)$. Put everything together the space cost for computing $\nabla_{\mathbf{U}}\mathcal{L}$ is $O(m+nd)$.

The time complexity of the part within $[\cdot]$ in Eq.\eqref{eq:explicit approximated hyper-gradient} is $O(nd^2+md)$. The time complexity of computing $\mathbf{W}^{\prime}\mathbf{X}^{\prime}\mathbf{V}$ is $O(ndr)$. The time complexity about the multiplication between $[\cdot]$ and $\mathbf{W}^{\prime}\mathbf{X}^{\prime}\mathbf{V}$ is $O(ndr)$. Hence, put everything together the total time complexity for computing $\nabla_{\mathbf{U}}\mathcal{L}$ is $O(nd^2+md)$ given $r<<d$.
\end{proof}

\section{Case study about the behaviour of \gasoline}

Here, we further study the potential reasons behind the success of \gasoline. To this end, we conduct a case study whose core idea is to label malicious modifications (from adversaries) and test if \gasoline\ is able to detect them. The specific procedure is that we utilize different kinds of attackers (i.e., metattack~\cite{DBLP:conf/iclr/ZugnerG19}, \nettack~\cite{zugner2018adversarial}, and random attack) to modify the graph structure of a \emph{benign} graph $G$ (with adjacency matrix $\mathbf{A}$) into a \emph{poisoned} graph $G_{\texttt{adv}}$ (with adjacency matrix $\mathbf{A}_{\texttt{adv}}$). Then, we utilize the score matrix $\mathbf{S}$ from Eq.~\eqref{eq:score matrix} to assign a score to every entry of the poisoned adjacency matrix $\mathbf{A}_{\texttt{adv}}$. As we mentioned in Section~\ref{sec:method}, the higher score an entry obtains, the more likely \gasoline\ will modify it. We compute the average score of three groups of entries from $\mathbf{A}_{\texttt{adv}}$: the poisoned entries after adding/deleting perturbations from adversaries, the benign existing edges without perturbation, and the benign non-existing edges without perturbation. Remark that both the benign graphs and the poisoned graphs are unweighted and we define following auxiliary matrices. $\mathbf{A}_{\texttt{diff}}=|\mathbf{A}_{\texttt{adv}}-\mathbf{A}|$ is a difference matrix whose entries with value $1$ indicate poisoned entries. $\mathbf{A}_{\texttt{benign-E}}=\mathbf{A}\odot(\mathbf{1}-\mathbf{A}_{\texttt{diff}})$ is a benign edge indicator matrix whose entries with value $1$ indicate the benign existing edges without perturbation. $\odot$ indicates element-wise multiplication. $\mathbf{A}_{\texttt{benign-NE}}=(\mathbf{1}-\mathbf{A})\odot(\mathbf{1}-\mathbf{A}_{\texttt{diff}})$ is a benign non-existing edge indicator matrix whose entries with value $1$ indicate the benign non-existing edges without perturbation. Based on that, we have the following three statistics:

\vspace{-4mm}
\begin{equation*}
\begin{split}
    S_{\texttt{adv}} &= \frac{\sum_{i,j} (\mathbf{S}\odot\mathbf{A}_{\texttt{diff}})[i,j]}{\sum_{i,j}\mathbf{A}_{\texttt{diff}}[i,j]},\\
    S_{\texttt{benign-E}} &= \frac{\sum_{i,j} (\mathbf{S}\odot\mathbf{A}_{\texttt{benign-E}})[i,j]}{\sum_{i,j}\mathbf{A}_{\texttt{benign-E}}[i,j]},\\
    S_{\texttt{benign-NE}} &= \frac{\sum_{i,j} (\mathbf{S}\odot\mathbf{A}_{\texttt{benign-NE}})[i,j]}{\sum_{i,j}\mathbf{A}_{\texttt{benign-NE}}[i,j]},\\
\end{split}
\end{equation*}
\noindent which denote the average score obtained by poisoned entries, benign existing edges, and benign non-existing edges.

\begin{figure}[t!]
\begin{subfigure}{.47\textwidth}
  \centering
  \includegraphics[width=.95\linewidth]{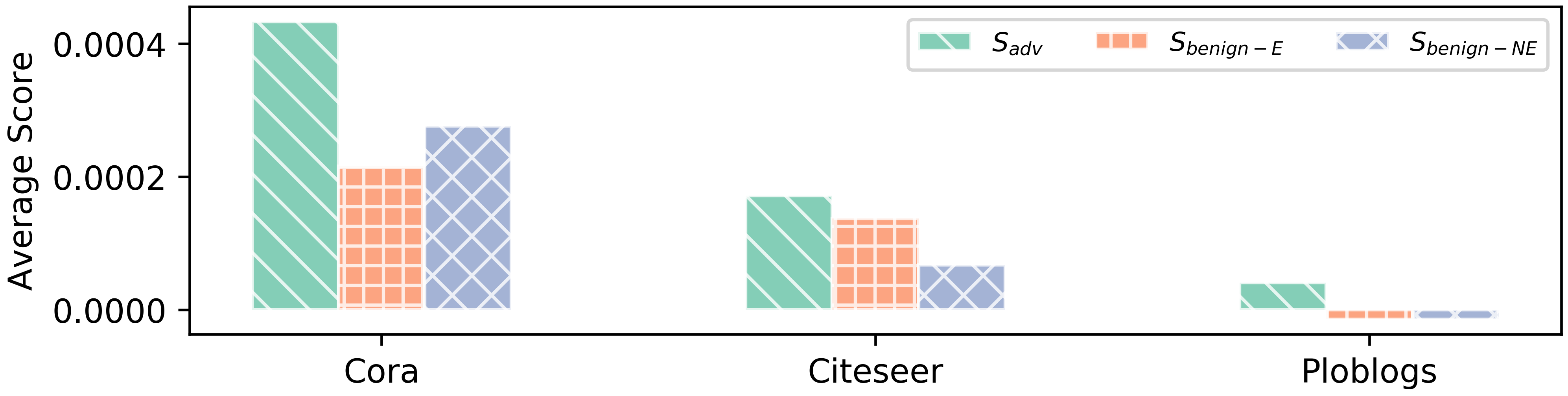}
  \vspace{-2mm}
  \caption{metattack}
\end{subfigure}
\begin{subfigure}{.47\textwidth}
  \centering
  \includegraphics[width=.95\linewidth]{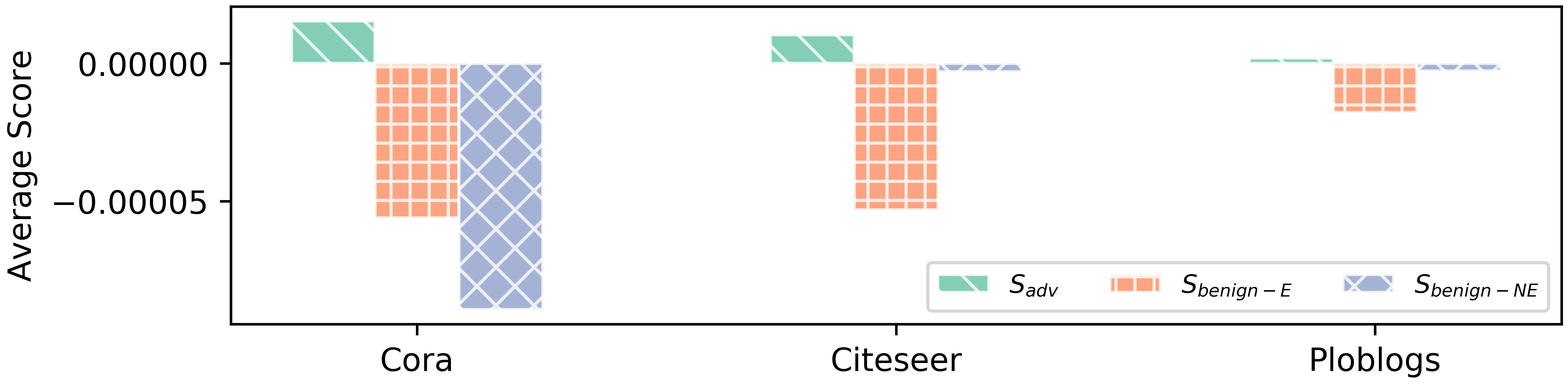}  
  \vspace{-2mm}
  \caption{\nettack}
\end{subfigure}
\begin{subfigure}{.47\textwidth}
  \centering
  \includegraphics[width=.95\linewidth]{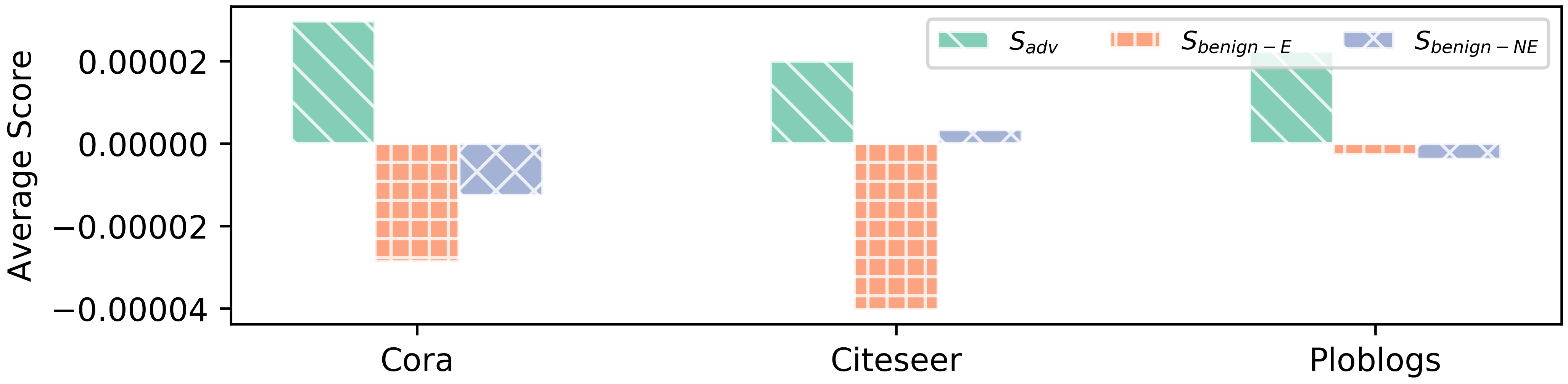}  
  \vspace{-2mm}
  \caption{random attack}
\end{subfigure}
\vspace{-4mm}
\caption{Score of various entries under metattack (a), \nettack~ (b), and random attack (c). Best viewed in color.}
\label{fig:case study}
\end{figure}

Detailed results are presented in Figure~\ref{fig:case study}. We observe that \gasoline~ tends to modify poisoned entries more (with higher scores) than to modify benign unperturbed entries in the adjacency matrix of poisoned graphs, which is consistent with our expectation and enables the algorithm to partially recover the benign graphs and to boost the performance of downstream classifiers.

\section{Effect of Modification Budget}
\label{sec:budget study}
\begin{figure}[h!]
\begin{subfigure}{.235\textwidth}
  \centering
  \includegraphics[width=1\linewidth]{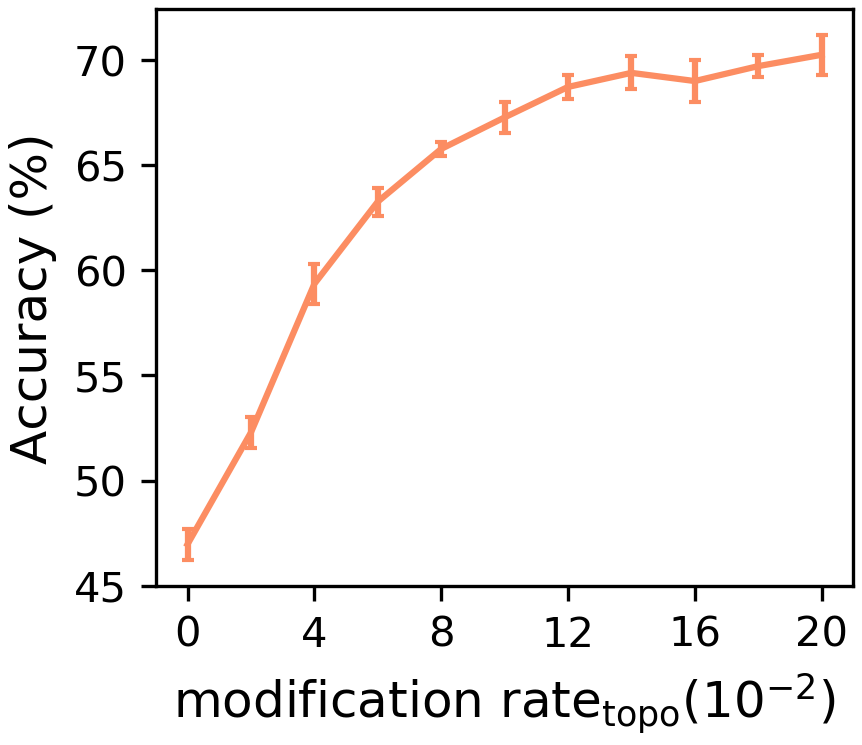}
  \vspace{-6mm}
  \caption{\gasoline-DT}
\end{subfigure}
\begin{subfigure}{.235\textwidth}
  \centering
  \includegraphics[width=1\linewidth]{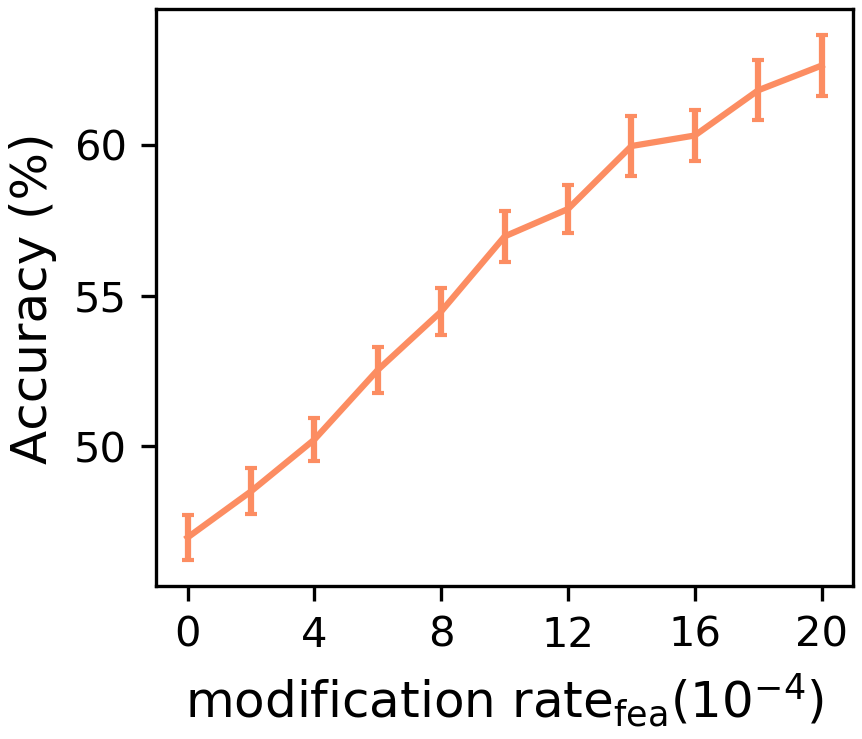}
  \vspace{-6mm}
  \caption{\gasoline-CF}
\end{subfigure}
\caption{Performance of downstream classifier vs. the modification budget of \gasoline-DT (a) and \gasoline-CF (b)}
\label{fig:budget study}
\end{figure}

In this section we study the relationships between the budget of \gasoline~ and the corresponding performance of the downstream classifier. Here, we instantiate two variants of \gasoline: discretized modification towards topology (\gasoline-DT) and continuous modification towards feature (\gasoline-CF). The provided graph is Cora~\cite{kipf2016semi} which is heavily-poisoned by metattack~\cite{DBLP:conf/iclr/ZugnerG19} with $\texttt{perturbation rate}=25\%$ (i.e., $B$). The perturbation budget per modification step $b$ is set to be $\frac{B}{10}$. Both the backbone classifier and the downstream classifier of \gasoline~ are the APPNP~\cite{klicpera2018predict} models with the aforementioned settings. From Figure~\ref{fig:budget study} we observe that with the increase of the modification budget ($\texttt{modification rate}_{\texttt{topo}}$ and $\texttt{modification rate}_{\texttt{fea}}$), \gasoline~enjoys great potential to further improve the performance of the downstream classifiers. At the same time, `economic' choices are strong enough to benefit downstream classifiers so we set $\texttt{modification rate}_{\texttt{topo}}$ as $0.1$ and $\texttt{modification rate}_{\texttt{fea}}$ as $0.001$ throughout our experiment settings.

\begin{figure}[h!]
\begin{subfigure}{.23\textwidth}
  \centering
  \includegraphics[width=\linewidth]{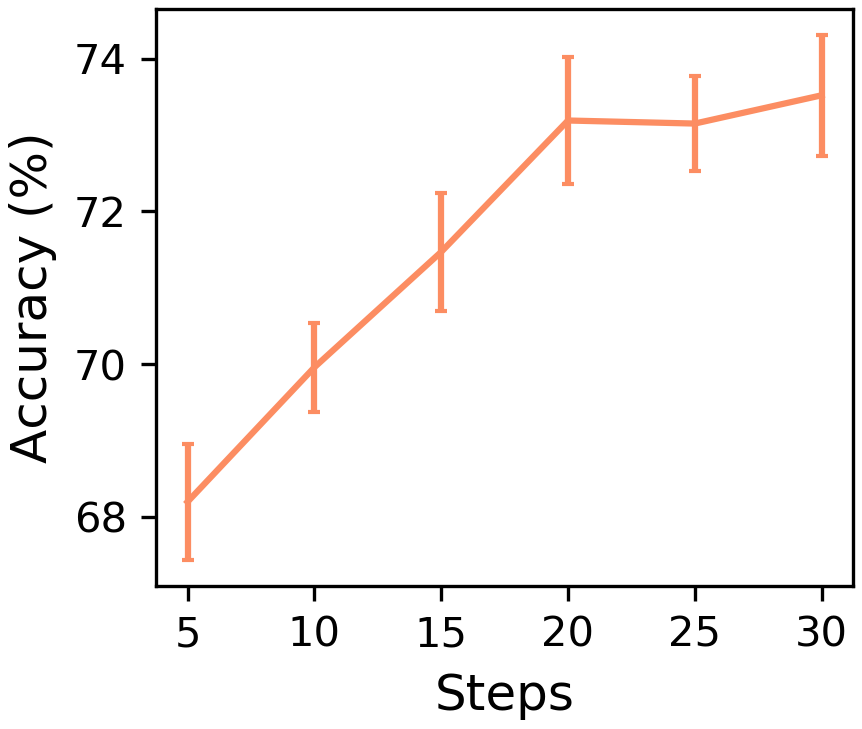}
  \vspace{-6mm}
  \caption{update steps}
  \label{fig:step study}
\end{subfigure}
\begin{subfigure}{.24\textwidth}
  \centering
  \includegraphics[width=\linewidth]{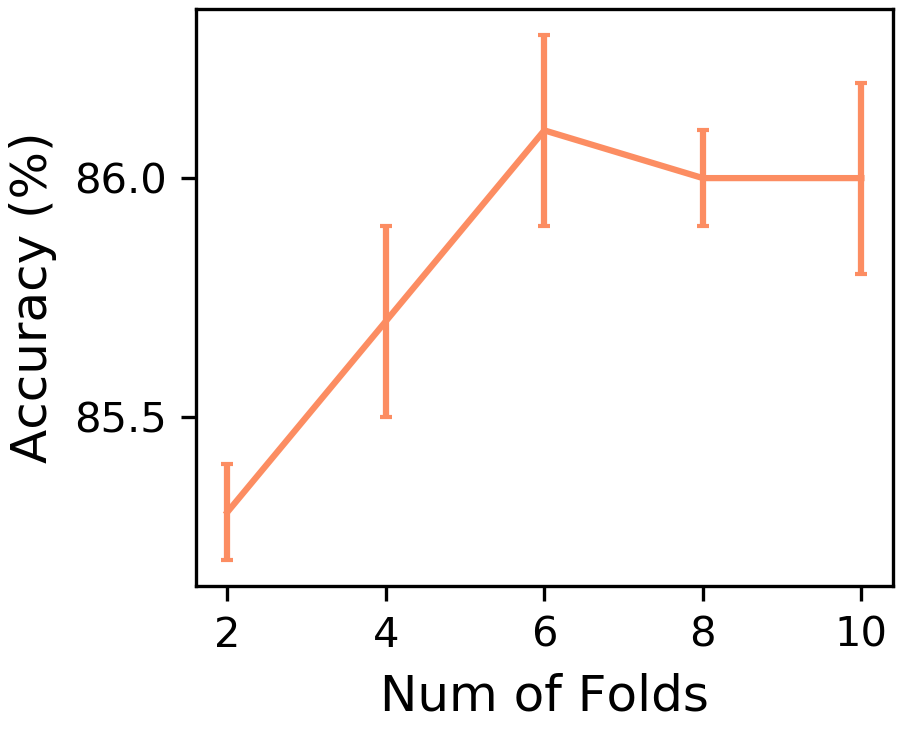}  
  \vspace{-6mm}
  \caption{number of folds $K$}
  \label{fig:folds study}
\end{subfigure}
\caption{Performance of \gasoline-DTCF vs. the update steps (a) and the number of folds $K$ (b).}
\end{figure}

\section{Effects of Modification Steps and Number of Folds}
As we claimed in the main content, in implementation we set the budget in every iteration as $b$ and update the given graph multiple iterations till we run out of total budget $B$. Hence, the update steps equals to $\lceil\frac{B}{b}\rceil$. Intuitively less budget per iteration can provide finer update towards the given graphs. To validate that we test the performance of an instantiation of \gasoline~ with discretized modification towards topology and continuous modification towards feature (\gasoline-DTCF) on the Cora~\cite{kipf2016semi} graph which is poisoned by metattack~\cite{DBLP:conf/iclr/ZugnerG19} with $\texttt{perturbation rate}=25\%$. Both the backbone classifier and the downstream classifier of \gasoline~ are the APPNP~\cite{klicpera2018predict} with the aforementioned settings. From Figure~\ref{fig:step study} we observe that with more update steps downstream classifiers can get better performance. However, when the number of steps is larger than $20$, the improvement of performance is minor.

In addition, the number of training/validation split fold $K$ is another important hyper-parameter in our model. Intuitively larger $K$ leads into better usage of the given data. To study the relationships between $K$ and the corresponding performance of the downstream classifier, we implement \gasoline-DTCF on the original Cora graph to verify that. Note that the $\texttt{modification rate}_\texttt{topo}=0.1$, $\texttt{modification rate}_\texttt{fea}=0.001$, and the number of modification steps is set as $10$. From Figure~\ref{fig:folds study} we observe that performance of the downstream classifier is improved with the increase of the number of folds. However, such performance gaining stops when $K=6$. Hence, $K=6$ is enough to make full use of the given graph by \gasoline.



\end{document}